\documentclass[12pt,a4paper]{article}

\usepackage[useregional]{datetime2}

\usepackage[british]{babel}

\usepackage[a4paper,top=2cm,bottom=2cm,left=2.5cm,right=2.5cm,marginparwidth=1.75cm]{geometry}

\usepackage[style=apa, backend=biber]{biblatex} 
\DeclareLanguageMapping{british}{british-apa} 
\DeclareFieldFormat[article]{volume}{\apanum{#1}} 

\DeclareSourcemap{
  \maps[datatype=bibtex]{
    \map[overwrite=true]{
       \pertype{misc}
       \step[fieldset=urldate,fieldvalue={\DTMtoday}]
    } 
 }
}

\usepackage{amsmath}
\usepackage{graphicx, adjustbox, xcolor}
\usepackage[colorlinks=true, allcolors=blue]{hyperref}
\usepackage{hyperref}
\usepackage{orcidlink}
\usepackage[title]{appendix}
\usepackage{mathrsfs}
\usepackage{amsfonts}
\usepackage{booktabs} 
\usepackage{caption}  
\usepackage{threeparttable} 
\usepackage{algorithm}
\usepackage{algorithmicx}
\usepackage{algpseudocode}
\usepackage{listings}
\usepackage{enumitem}
\usepackage{chngcntr}
\usepackage{booktabs}
\usepackage{lipsum}
\usepackage{subcaption}
\usepackage{authblk}
\usepackage[T1]{fontenc}    
\usepackage{csquotes}       
\usepackage{diagbox}
\usepackage{comment}

\usepackage{makecell}
\usepackage{tabularray}

\algrenewcommand\algorithmicrequire{\textbf{Input:}}
\algrenewcommand\algorithmicensure{\textbf{Output:}}
\DeclareMathOperator*{\argmin}{arg\,min}
\DeclareMathOperator*{\argmax}{arg\,max}

\usepackage{subcaption}
\usepackage[justification=raggedright, singlelinecheck=false]{caption}
\usepackage{caption}
\usepackage{helvet}  
\usepackage{lmodern}  

\usepackage{setspace}
\usepackage{titlesec}
\titleformat{\section} 
  {\normalfont\Large\bfseries}{\thesection.}{1em}{}
  
\usepackage{lineno} 

\rightlinenumbers 

\usepackage{float}   
\usepackage{caption} 


\title{Airfoil optimization using Design-by-Morphing with minimized design-space dimensionality}

\author[1]{Sangjoon Lee}
\author[2,$\,$*]{Haris Moazam Sheikh}
\affil[1]{\small Center for Turbulence Research, Stanford University, Stanford, CA 94305, USA, https://orcid.org/0000-0002-2063-6298}
\affil[2]{\small Department of Aeronautics and Astronautics, University of Southampton, Southampton SO17 1BJ, UK, https://orcid.org/0000-0002-3154-0494}
\affil[*]{\small Corresponding author. \texttt{h.m.sheikh@soton.ac.uk}}

\date{}  

\begin{document}
\maketitle

\begin{abstract}
Effective airfoil geometry optimization requires exploring a diverse range of designs using as few design variables as possible. This study introduces AirDbM, a Design-by-Morphing (DbM) approach specialized for airfoil optimization that systematically reduces design-space dimensionality. AirDbM selects an optimal set of 12 baseline airfoils from the UIUC airfoil database, which contains over 1,600 shapes, by sequentially adding the baseline that most increases the design capacity. With these baselines, AirDbM reconstructs 99 \% of the database with a mean absolute error below 0.005, which matches the performance of a previous DbM approach that used more baselines. In multi-objective aerodynamic optimization, AirDbM demonstrates rapid convergence and achieves a Pareto front with a greater hypervolume than that of the previous larger-baseline study, where new Pareto-optimal solutions are discovered with enhanced lift-to-drag ratios at moderate stall tolerances. Furthermore, AirDbM demonstrates outstanding adaptability for reinforcement learning (RL) agents in generating airfoil geometry when compared to conventional airfoil parameterization methods, implying the broader potential of DbM in machine learning-driven design.
\end{abstract}


\section*{Nomenclature}
\subsection*{Alphabets and greek letters}
\begin{tabbing}
$\mathcal{A}$\qquad \= Arbitrary airfoil shape \\
$\mathcal{B}_i$\qquad \= $i$-th baseline airfoil shape \\
$c$\qquad \= Airfoil chord length ($\mathrm{m}$) \\
$c_s$\qquad \= Speed of sound ($\mathrm{m~s^{-1}}$) \\
$d$\qquad \= Drag force exerted on an airfoil per unit span ($\mathrm{N~m^{-1}}$) \\
$l$\qquad \= Lift force exerted on an airfoil per unit span ($\mathrm{N~m^{-1}}$) \\
$\mathcal{M}$\qquad \= Morphed airfoil shape \\
$N$\qquad \= Design-by-Morphing shape normalization factor \\
$S$\qquad \= Airfoil shape similarity measure \\
$U$\qquad \= Freestream flow speed ($\mathrm{m~s^{-1}}$) \\
$w_i$\qquad \= Design-by-Morphing weight factor with respect to the $i$-th baseline \\
$x$\qquad \= Horizontal Cartesian coordinate \\
$y$\qquad \= Vertical Cartesian coordinate \\
$\alpha$\qquad \= Airfoil angle of attack (${}^{\circ}$) \\
$\alpha_s$\qquad \= Airfoil stall angle, the first local maximum of $\alpha$ with respect to $l$ (${}^{\circ}$) \\
$\Delta \alpha$\qquad \= Stall tolerance, the range of $\alpha$ between $\alpha_s$ and the maximum $l/d$ point (${}^{\circ}$) \\
$\nu$\qquad \= Fluid kinematic viscosity ($\mathrm{m^2~s^{-1}}$) \\
$\rho$\qquad \= Fluid density ($\mathrm{kg~m^{-3}}$) \\
\end{tabbing}

\subsection*{Dimensionless groups}
\begin{tabbing}
$C_d$\qquad \= Drag coefficient, $2d/(\rho U^2 c)$ \\
$C_l$\qquad \= Lift coefficient, $2l/(\rho U^2 c)$ \\
$\mathrm{Re}$\qquad \= Reynolds number based on airfoil chord length, $Uc/\nu$ \\
$\mathrm{Ma}$\qquad \= Mach number, $U/c_s$
\end{tabbing}

\section{Introduction}\label{introduction}

In aerodynamic design, airfoil shape optimization remains a fundamental and active challenge that requires the exploration of a design space comprising diverse and valid airfoil configurations \parencite{Drela1998, Lyu2015, Skinner2018, Martins2022}. Evaluating the performance of each design involves analyzing multiple dynamic metrics (e.g., lift, drag, and stall angle) across varying flight conditions (e.g., wind speed and angle of attack), often demanding computationally intensive simulations to identify optimal candidates. As in typical optimization processes, the first step is to define a design space that captures a broad range of airfoil shapes using a finite set of parameters, enabling systematic exploration and refinement toward optimal solutions \parencite{Sobester2008, Masters2017}.

A number of studies have explored airfoil shape parameterization methods, including --- but not limited to --- PARSEC \parencite{Sobieczky1999}, Class-Shape Transformation (CST) \parencite{Kulfan2006}, Hicks-Henne bump functions \parencite{Hicks1978}, Bézier curves \parencite{Gordon1974, Derksen2010} or Non-Uniform Rational B-Splines (NURBS) \parencite{Piegl1997, Lepine2000}, which form the basis of Free-Form Deformation (FFD) \parencite{Sederberg1986}. Recently, with advances in machine learning, deep generative models have also been explored for airfoil shape parameterization via nonlinear dimensionality reduction \parencite{Chen2020, Xie2024}. Each method provides a systematic approach to achieving design flexibility based on distinct mathematical principles. The choice of parameterization significantly affects the diversity of airfoil shapes within the constructed design space. If the global design space of airfoil shapes (representing the maximum possible diversity) were known, the optimal method would be the one that constructs this space with the fewest parameters, thereby mitigating the \textit{curse of dimensionality} \parencite{Sobester2008, Viswanath2011, Serani2024}.

However, in a general sense, the \textit{maximum} level of design diversity is hardly attainable. Many aerodynamic and hydrodynamic shape design problems --- such as high-speed train aerodynamics \parencite{Oh2018}, riblet surface design for drag reduction \parencite{Bai2016, Lee2024}, or hydrokinetic turbine draft tube optimization \parencite{Sheikh2022} --- even suffer from a lack of design diversity for several reasons. First of all, these problems inherently involve highly nonlinear dynamics in design evaluation, making the correlation between geometry and performance non-intuitive and difficult to predict. Additionally, many of the practical or commercially adopted designs, often developed through costly trial-and-error processes (because of the first reason), remain proprietary and are not publicly available \parencite{Benjamin2025}. As a result, design exploration in such problems is significantly restricted by a lack of a rich, comprehensive, and centralized design database, leaving room for novel designs that have yet to be explored.

In this respect, Design-by-Morphing (DbM) has been proposed to offer a universal strategy to these challenges across different design problems by enabling extensive design space exploration based on a limited set of baseline designs \parencite{Oh2018, Sheikh2022, Sheikh2023, Lee2024}. Rather than relying on predefined shape parameterizations which are mostly problem-specific (e.g., PARSEC for airfoils), DbM generates intermediate forms by morphing between selected baselines, facilitating the constraint-free and continuous creation of new designs. The weight factors assigned to the baselines replace traditional shape parameters, meaning that the number of baselines determines the dimensionality of the design space. DbM also allows for \textit{extrapolative} morphing, wherein negative weight factors can be applied to some or all baselines while handling non-feasible geometries (e.g., self-intersections), which increases the ability to encompass novel shapes to expand the design space beyond the interpolative morphing of the baseline design set. Another major advantage of this framework comes from its inherent interpretability; the weight factors explicitly quantify the geometric influence of each baseline airfoil, enabling intuitive design adjustments that are directly comprehensible.

Although DbM offers clear theoretical advantages in design space exploration, it is important to rigorously assess its practical effectiveness in comparison to established shape parameterization methods. The airfoil design and optimization problem serves as an ideal benchmark for this purpose, given the availability of several conventional airfoil shape parameterization techniques that provide meaningful standards for comparison. Not only that, unlike many other aerodynamic and hydrodynamic shape design problems, but airfoil design also benefits from a rich, publicly available database containing more than 1,600 airfoil shapes as of now \parencite{SeligUIUC}. This extensive repository (hereinafter denoted as the UIUC database), a result of over a century of modern airfoil development \parencite{Bilstein1989, Anderson1997}, offers a diverse set of tested and proposed design alternatives. While the UIUC database may not perfectly represent the global airfoil design space, its breadth and diversity make it sufficiently comprehensive to be regarded as approximately global for practical purposes.

From this perspective, DbM for airfoil optimization was evaluated in the authors’ previous study \parencite{Sheikh2023}. DbM was applied to reconstruct the entire UIUC database, with its performance compared against PARSEC, NURBS, and the Hicks-Henne approach. The results confirmed DbM’s competitiveness in airfoil shape generation and demonstrated the importance of extrapolative morphing in expanding the design space. However, in that study, the baseline selection of 25 airfoil shapes relied on designer judgment, raising questions about DbM’s sensitivity to baseline selection (though partial mitigation was achieved through subset analysis, variations across distinct baseline sets remained unexamined). Such manual curation risks unintended biases in design-space coverage. Note that this concern arises specifically when a diverse, representative baseline dataset exists (e.g., more than hundreds of design points); in contexts lacking comprehensive design sets (less than 10 design points), all existing designs can simply be used as baselines for DbM \parencite[e.g.,][]{Sheikh2022}.

Given the presence of comprehensive accessible designs, if DbM is able to achieve the same design generation capacity with their small subset, it can further benefit optimization by reducing the dimensionality of the design space, resulting in more time-efficient optimum search and faster convergence. We address this issue by introducing a systematic baseline identification process that mitigates designer bias while maximizing design-space representativeness. Focusing on a specific case of DbM for airfoil design and optimization, the current study aims to provide the following contributions:
\begin{itemize}
    \item Developing an effective approach to identify reduced baseline sets for DbM while maintaining its airfoil design generation capacity.
    \item Presenting an optimal baseline set for DbM with reduced design-space dimensionality, which rivals the precedent with a larger number of airfoil baselines.
    \item Quantifying improvements in airfoil shape design and optimization using a new DbM with reduced design-space dimensionality.
\end{itemize}

We first briefly revisit the application of DbM in 2D airfoil design, describing the details of morphing, the similarity measure between airfoil shapes, and the airfoil reconstruction problem with additional clarifications from our previous study. Several approaches for identifying optimal baseline selections (assuming the global design space is represented by the UIUC database) are then discussed, along with an analysis of which approach is the most feasible given limited computing resources. Next, using the baseline set revealed through this approach, example cases of airfoil optimization are conducted with the objectives of maximizing the lift-to-drag ratio and stall angle tolerance, quantifying convergence acceleration and solution enhancements. Finally, we demonstrate DbM's adaptability in reinforcement learning environments for airfoil geometry generation, enabling designers to achieve faster learning rates and higher accuracy than conventional airfoil parameterization methods.


\section{Design-by-Morphing for Airfoil Optimization}\label{dbmaf}

\begin{figure}[t!]
    \centering
    \includegraphics[width=\linewidth]{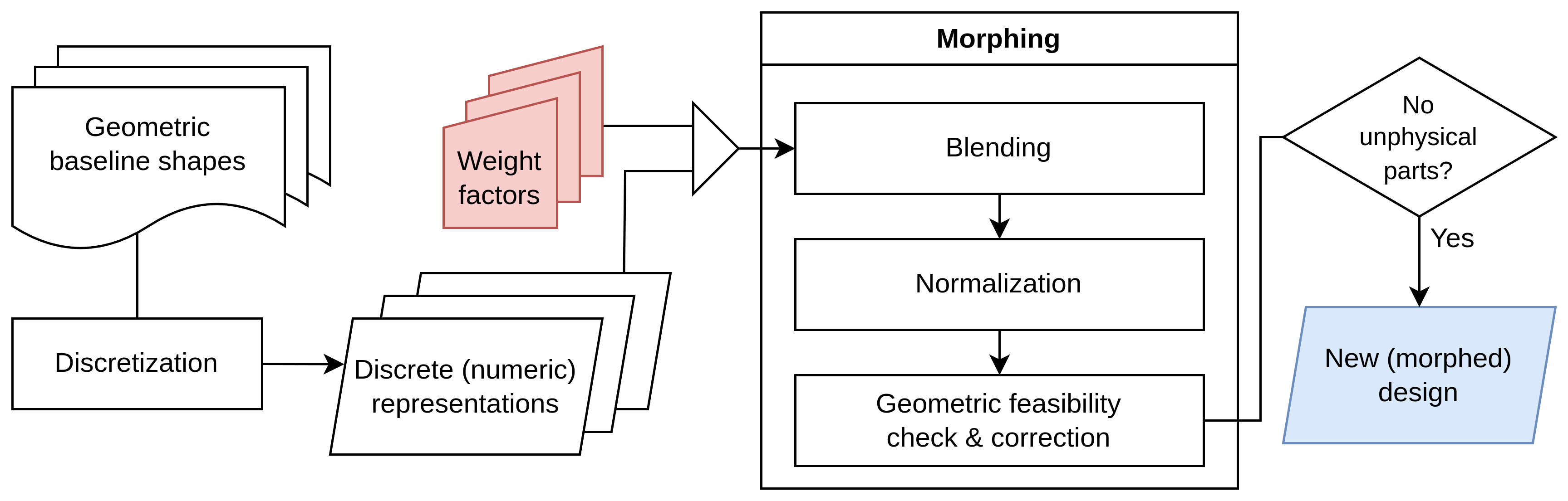}
    \caption{General flowchart of DbM to get a new design by \textit{morphing} baseline shapes.}
    \label{fig:DbMGeneralFlowchart}
\end{figure}

\subsection{Formulation for airfoil morphing}
In this section, we review and summarize the 2D airfoil design process using DbM, which was described in \textcite{Sheikh2023}. While the core procedure remains identical, we provide additional details that were not thoroughly covered in the previous study to enhance the robustness of this design technique. For this purpose, we first outline the general steps of DbM before applying them specifically to airfoil morphing.

Figure \ref{fig:DbMGeneralFlowchart} presents a generalized flowchart of DbM, outlining the sub-processes involved in outputting a newly morphed design from selected baselines, where the user specifies weight factors for each baseline as input. This flowchart assumes that the baseline designs have already been selected; the reduction of design-space dimensionality (i.e., using fewer baselines) is not the focus here but will be addressed later (\S \ref{baselinesel}).

In the pre-morphing stage, baseline shapes, originally defined in geometric form, must be converted into a discrete numerical representation in a consistent format to enable basic arithmetic operations (e.g., addition and scalar multiplication) for computational processing. This concept is widely studied in computer animation for object transformation, where various techniques have been developed \parencite{Parent2012}. We refer to this process as \textit{discretization} to emphasize the mapping of geometric shapes into a consistent numerical form, such as control points or grids.

Next, the morphing stage consists of three sub-processes: (1) \textit{blending}, where the baseline shapes are combined according to the input weight factors; (2) \textit{normalization}, which scales the blended shape to fit within typical scale of the problem; and (3) \textit{geometric feasibility check and correction}, which adjusts and removes any unphysical parts, {\color{black} mostly represented by self-intersections}. If no unphysical parts remain in the final shape, the process is complete, and a new design is achieved. Notably, the overall process resembles the metamorphosis of irregularly shaped (e.g., non-rectangular) objects in computer graphics and, in 2D, several practical approaches to these general procedures have been considered and developed \parencite{Sederberg1992, vandenBergh2002}.

\begin{figure}[t!]
    \centering
    \includegraphics[width=.75\linewidth]{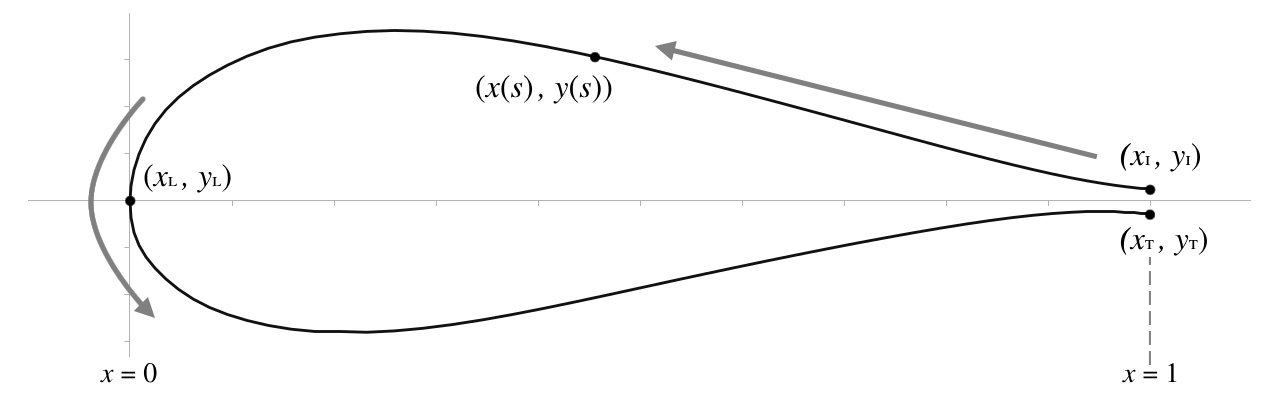}
    \caption{Selig coordinate format for airfoil geometry of a unit chord length \parencite[see][]{SeligUIUC}.}
    \label{fig:SeligFormat}
\end{figure}

When it comes to airfoil morphing, one of the most widely used formats for describing airfoil geometry is the Selig coordinate format, or simply the Selig format. Named after Selig, this format is used to store airfoil data in a structured manner. As shown in Figure \ref{fig:SeligFormat}, it consists of a list of ($x$, $y$) coordinate pairs that define the airfoil geometry non-dimensionalized by chord length $c$. The coordinates are arranged sequentially, starting from the upper trailing edge ($x=x_I=1$), following the upper surface toward the leading edge ($x=x_L=0$), and then continuing along the lower surface back to the lower trailing edge ($x=x_T=1$). Using this format, any arbitrary airfoil shape $\mathcal{A}$ can be represented as a parametric curve with respect to a variable $s$, defined as:
\begin{equation}
    \mathcal{A}(s) \equiv \begin{pmatrix} x(s), ~ y(s) \end{pmatrix}~~~0\le s \le 2 ,
    \label{eq:airfoilcurveparameterize}
\end{equation}
where $x(s) \equiv |1 - s|$ and $y(s)$ depends on the specific airfoil geometry (which thereby defines it). Discretizting $s$ into equispaced points $s_j$ for $j = 0,~1,~\cdots,~F$, such that $0 = s_0<s_1<\cdots<s_F = 2$ with $s_j = 2j/F$, we obtain $\mathcal{A}$'s discrete numerical representation as the following $(F+1)$-dimensional vector (referred to as the Selig-format vector henceforth):
\begin{equation}
    \vec{\mathcal{A}} \equiv \begin{bmatrix}~y(s_0) & y(s_1) & \cdots ~ y(s_F)~\end{bmatrix}^T \in \mathbb{R}^{F+1} .
    \label{eq:airfoildiscretize}
\end{equation}
We assume that $F$ is sufficiently large so that the airfoil’s shape is well preserved, with minimal loss of geometric detail between consecutive discretized points. In practice, $F=200$ is found to be large enough to represent every airfoil in the UIUC database. In Equation \ref{eq:airfoildiscretize}, $y(s_0) = y(0) = y_I$, $y(s_{F/2}) = y(1) = y_L$, and $y(s_F) = y(2) = y_T$ (in order for $F/2$ to be integer, let's assume $F$ to be even).

Given $n$ baseline airfoil shapes $\mathcal{B}_1$, $\mathcal{B}_2$, $\cdots$ $\mathcal{B}_n$, each can be expressed in Selig-format vector form as $\vec{\mathcal{B}}_1$, $\vec{\mathcal{B}}_2$, $\cdots$ $\vec{\mathcal{B}}_n$ in $\mathbb{R}^{F+1}$. Since these baseline shapes correspond one-to-one and $\mathbb{R}^{F+1}$ is a vector space that is equipped with well-defined addition and scalar multiplication, we can formally define the process of airfoil morphing. Topologically, the existence of a one-to-one mapping is ensured by the homeomorphism of these shapes, which is a prerequisite for performing DbM, as noted in \textcite{Sheikh2023}. Our previous study highlighted the homeomorphism of 2D closed shapes. However, in consideration of the fact that the geometric representation an airfoil here is a closed curve as in Equation \ref{eq:airfoilcurveparameterize}, the relevant homeomorphism to be correctly highlighted should be that of a 1-manifold with boundary, topologically equivalent to closed intervals. For mathematical rigor, in the case of airfoils with zero trailing edge thickness ($y(0) - y(2) = 0$), let zero be interpreted as \textit{almost zero} ($y(0) - y(2) = 0^+$) to preserve the same homeomorphism (by conceptually separating the endpoints).

Overall, morphing of the $n$ baseline airfoil shapes with given weight factors $w_1$, $w_2$, $\cdots$, $w_n$ $\in [-1,1]$ is expressed as follows:
\begin{equation}
    \vec{\mathcal{M}} = \mathcal{F} \left( \frac{1}{N}\sum_{i=1}^{n}{w_i \vec{\mathcal{B}_i}} \right) ,
    \label{eq:morphing}
\end{equation}
where $N = N(w_1,w_2,\cdots,w_n)$ is a normalization factor that scales the blended shape, and $\mathcal{F}$ represents a set of adjustment operations to correct unphysical geometries. A linear blending formula is chosen as it represents the simplest form of blending. However, the choice of blending is not necessarily limited to linear methods, and users may also explore nonlinear blending approaches. Similarly, the normalization factor can be determined in various ways, but we adhere to the original formulation:
\begin{equation}
    N = \sum_{i=1}^{n}w_i ,
    \label{eq:morphingnorm}
\end{equation}
to maintain consistency with the original study for comparative purposes. As for checking and correcting geometric feasibility, the removal of self-intersections is essential. Additional treatments may be applied depending on what design features users consider unphysical (e.g., holes). In the present airfoil morphing process, we focus solely on treating self-intersections. If no self-intersections are present, $\mathcal{F}$ is the identity map. Otherwise, in general, we adopt the self-intersection removal procedure for airfoil shapes as introduced in \S 2.2 of \textcite{Sheikh2023}, in which zero-thickness points due to self-intersections are locally stiffened and then smoothed. Starting with the Selig-format vector representation, we can inexpensively detect self-intersections in an airfoil shape using a simple sign-change checker. The detection algorithm is presented in Algorithm \ref{alg:detectselfintersections}.

\begin{algorithm}[!tb]
\caption{Detect an self-intersecting airfoil shape given as a Selig-format vector}\label{alg:detectselfintersections}
\begin{algorithmic}
\Require $\vec{\mathcal{A}} = \left[~ y(s_0) ~~ y(s_1) ~~ \cdots ~~ y(s_F) ~\right]^T$ where $F$ is even (e.g., $F= 200$)
\Ensure \texttt{True} if $\vec{\mathcal{A}}$ represents a self-intersecting shape, else \texttt{False} 
\State $\vec{\mathcal{A}}_u \Leftarrow  [~ y(s_{F/2-1}) ~~ \cdots ~~ y(s_0) ~]^T$ \hfill {\footnotesize // slice the first half of $\vec{\mathcal{A}}$, and flip it}
\State $\vec{\mathcal{A}}_l \Leftarrow  [~ y(s_{F/2+1}) ~~ \cdots ~~ y(s_{F}) ~]^T$ \hfill {\footnotesize // slice the second half of $\vec{\mathcal{A}}$}
\State Allocate $\vec{b}$ (integer array) of size $F/2$
\For{\texttt{i} $\Leftarrow$ $1$ to $F/2$} \hfill {\footnotesize // one-based indexing is assumed}
    \State $x$ $\Leftarrow$ $\vec{\mathcal{A}}_u$\texttt{[i]} - $\vec{\mathcal{A}}_l$\texttt{[i]} 
    \State $\vec{b}\texttt{[i]}$ $\Leftarrow$ Sign of $x$ (1 for positive values, -1 for negative values, and 0 for zero)  
\EndFor
\For{\texttt{j} $\Leftarrow$ $1$ to $F/2-1$}
    \If{($\vec{b}$\texttt{[j]} \texttt{*} $\vec{b}$\texttt{[j+1]} $<$ $0$) \texttt{.OR.} ($\vec{b}$\texttt{[j]} \texttt{==} $0$)}
        \State \Return \texttt{True}
    \EndIf
\EndFor
\State \Return \texttt{False}
\end{algorithmic}
\end{algorithm}

\subsection{Shape similarity measure}
For two arbitrary airfoil shape geometries $\mathcal{A}_1$ and $\mathcal{A}_2$ in the parametric curve form with respect to $s$ as in Equation \ref{eq:airfoilcurveparameterize}, their similarity, denoted as $S (\mathcal{A}_1, \mathcal{A}_2 )$, can be quantified by measuring the mean absolute error between these two airfoils along the upper and lower surfaces, respectively, and then summing the results. That is,
\begin{equation}
    S (\mathcal{A}_1, \mathcal{A}_2 ) \equiv \underbrace{\frac{\int_0^1 | y_1 (s) - y_2 (s)|ds}{\int_0^1{ds}}}_{\text{Upper curve}} + \underbrace{\frac{\int_1^2 | y_1 (s) - y_2 (s)|ds}{\int_1^2{ds}}}_{\text{Lower curve}} ,
    \label{eq:similarity_1}
\end{equation}
where $y_1 (s)$ and $y_2 (s)$ are the $y$-coordinates of $\mathcal{A}_1 (s)$ and $\mathcal{A}_2 (s)$, respectively. Since both $\int_0^1 ds$ and $\int_1^2 ds$ evaluate to unity, Equation \ref{eq:similarity_1} simplifies to
\begin{equation}
    S (\mathcal{A}_1, \mathcal{A}_2 ) = \int_0^2 | y_1 (s) - y_2 (s)|ds .
    \label{eq:similarity_2}
\end{equation}
Here, $\int_0^2{(y_1 (s) -y_2 (s))ds = 0}$ is assumed to provide a consistent vertical alignment. This formulation is equivalent to the airfoil shape similarity measure (as mean absolute error) proposed by \textcite{Sheikh2023}. It is important to note that various similarity measures can be defined as long as they form a convergent series in which the similarity value approaches a certain limit (mostly zero) as one shape becomes identical to the other; Equation \ref{eq:similarity_2} evidently satisfies this fundamental requirement.

Taking one step further from \textcite{Sheikh2023}, let us derive a discretized formula that is effectively equivalent to Equation \ref{eq:similarity_2}. Considering an equispaced $(F+1)$-point discretization of $s$, we use numerical integration based on the trapezoidal rule to obtain an approximate form of Equation \ref{eq:similarity_2}, given by:
\begin{equation}
\begin{aligned}
    S (\mathcal{A}_1, \mathcal{A}_2 ) ~\simeq~ & \frac{2}{F} \sum_{i=1}^{F} \frac{1}{2} \left( | y_1 (s_{i-1} ) - y_2 (s_{i-1} )| + | y_1 (s_i ) - y_2 (s_i )| \right)
    \\ ~=~ & \sum_{i=1}^{F-1} \frac{2}{F} | y_1 (s_{i} ) - y_2 (s_{i} )| \\ & +~ \frac{1}{F} \left( | y_1 (s_{0} ) - y_2 (s_{0} )| + | y_1 (s_{F} ) - y_2 (s_{F} )| \right) .
    \label{eq:similarity_3}
\end{aligned}
\end{equation}
For a more compact expression, we may factor the endpoint terms into the summation by multiplying them by 2, resulting in:
\begin{equation}\
    S' (\mathcal{A}_1, \mathcal{A}_2 ) =\frac{2}{F} \sum_{i=0}^{F} | y_1 (s_{i} ) - y_2 (s_{i} )| .
    \label{eq:similarity_4}
\end{equation}
$S'$ can be interpreted as a modified version of $S$ that places slightly more weight on endpoint error evaluation. In the context of airfoils, this adjustment emphasizes matching the trailing edge, which can be rationalized since airfoil dynamics are considerably influenced by edge geometries. The right-hand side of Equation \ref{eq:similarity_4} corresponds to the mean absolute difference between the Selig-format vectors of $\mathcal{A}_1$ and $\mathcal{A}_2$ (scaled by $2(F+1)/F$, or approximately 2 when $F$ is much greater than 1). Using the $\ell^1$-norm notation, we express the similarity measure for Selig-format vectors as
\begin{equation}\
    S' (\vec{\mathcal{A}}_1, \vec{\mathcal{A}}_2 ) =\frac{2}{F} \left\lVert \vec{\mathcal{A}}_1 - \vec{\mathcal{A}}_2\right\rVert_1 ,
    \label{eq:similarity_5}
\end{equation}
which we use as the airfoil shape similarity measure that is effectively equivalent to the integral form in Equation \ref{eq:similarity_2}.

\subsection{Airfoil reconstruction problem}
Suppose that we aim to reconstruct a known airfoil shape, $\vec{\mathcal{A}_t}$, using the DbM process with given $n$ baseline shapes $\vec{\mathcal{B}}_1$, $\vec{\mathcal{B}}_2$, $\cdots$ $\vec{\mathcal{B}}_n$. It is additionally assumed that $\vec{\mathcal{A}_t}$ is distinct from each $\vec{\mathcal{B}}_i$ for any $i=1,~2,~\cdots,~n$; otherwise, the reconstruction is trivial. Recalling Figure \ref{fig:DbMGeneralFlowchart}, DbM employs $n$ input morphing weight factors, $w_1$, $w_2$, $\cdots$, $w_n$, to output a morphed airfoil shape $\vec{\mathcal{M}}$ as shown in Equation \ref{eq:morphing}. Defining the weight vector $\vec{w} \equiv [w_1~w_2~\cdots~w_n]^T \in \mathbb{R}^{n}$, we can formulate the problem of finding $\vec{w}$ as a single-objective optimization problem with continuous variables in standard form:
\begin{equation}
\begin{aligned}
    \argmin_{\vec{w} \; \in \;\mathbb{R}^n} \quad S' ( \vec{\mathcal{M}}(\vec{w}), \ \vec{\mathcal{A}_t}) \quad 
    \textrm{subject to} \quad \left\lVert \vec{w} \right\rVert_\infty \le 1 ,
    \label{eq:reconstructionstandardopt}
\end{aligned}
\end{equation}
where $\lVert \cdot \rVert_\infty$ represents the $\ell^\infty$-norm.

If there exists a set of $m$ airfoil shapes to be reconstructed, denoted as $\vec{\mathcal{A}}_{t,i}$ for $i = 1,~\cdots,~m$, solving the optimization problem in Equation \ref{eq:reconstructionstandardopt} $m$ times for each $\vec{\mathcal{A}}_{t,i}$ yields $m$ weight vectors $\vec{w}_{\text{opt}, i}$. These weight vectors generate morphed shapes that best approximate their respective target airfoil shapes. In each case, the proximity of $S'(\vec{\mathcal{M}}(\vec{w}_{\text{opt},i}), \vec{\mathcal{A}}_{t,i})$ to zero indicates how accurately the morphing of the $n$ baseline shapes reconstructs $\vec{\mathcal{A}}_{t,i}$. Consequently, the sum of these similarity measures, denoted as $S^\ddagger$, i.e., 
\begin{equation}
    S^\ddagger \equiv \sum_{i=1}^{m} S'(\vec{\mathcal{M}}(\vec{w}_{\text{opt},i}), \vec{\mathcal{A}}_{t,i}) ,
    \label{eq:sumofsimiliarities}
\end{equation}
can serve as an indicator of the reconstruction capability of the set of airfoil shape baselines, $\vec{\mathcal{B}}_1$, $\vec{\mathcal{B}}_2$, $\cdots$ $\vec{\mathcal{B}}_n$, for the target airfoil set, $\vec{\mathcal{A}}_{t,1}$, $\vec{\mathcal{A}}_{t,2}$, $\cdots$ $\vec{\mathcal{A}}_{t,m}$.

\section{Baseline Shape Selection}\label{baselinesel}
The central challenge addressed in this section --- and a pivotal question for this study --- is: How can we identify an optimal minimal set of baseline airfoil shapes that effectively represents the diversity of a larger collection considered globally representative? Selecting such a compact baseline set is conceptually analogous to principal component analysis (PCA), where lower-dimensional subspaces capture significant data variations \parencite[see][exhibiting the use of PCA to the UIUC database]{Li2022}. In our context, the selected baseline airfoils serve a role similar to principal components, enabling reconstruction of diverse airfoil designs through weighted morphing.

However, DbM's baseline selection process is distinct from PCA. While PCA generates abstract principal components through linear combinations, DbM preserves original baseline airfoils as interpretable building blocks. This approach maintains physical intuition by allowing designers to work with recognizable geometries rather than abstract eigencomponents whose meaning is obscured by PCA's rotational transformations. Furthermore, DbM introduces essential non-linearities through geometric feasibility corrections. Although linear blending is employed in this study (as in Equation \ref{eq:morphing}), the method permits extension to non-linear blending strategies, necessitating non-linear dimensionality reduction approaches unlike PCA. This structural flexibility is to enhance DbM's design generation capabilities while preserving geometric interpretability.

With these distinctions established, our objective is to identify a minimal subset from the global airfoil set $\left\{ \mathcal{A}_1, \cdots,\mathcal{A}_m \right\}$ that minimizes the total reconstruction error $S^{\ddagger}$ Equation \ref{eq:sumofsimiliarities} when used as DbM baselines. This maximizes the representational power of a compact design space while maintaining DbM's core advantages of physical interpretability and constrained dimensionality.

\subsection{Description of problem}\label{baselineselsub1}
To improve the practicality of DbM for airfoil design and optimization, we identify a set of baseline shapes that can effectively span the diversity of possible airfoil geometries. In order to define and quantify the coverage of the airfoil design space, it is necessary to first establish a comprehensive target set that reasonably represents the global airfoil shape design space. 

The UIUC airfoil database \parencite{SeligUIUC} provides a broad and diverse repository of airfoil geometries, accumulated through more than a century of aerodynamic development. Given its scope, historical depth, and inclusion of a multitude of tested, proposed, and optimized designs, the UIUC database can be regarded as approximately global for practical purposes (while no finite database can perfectly capture the infinite possibilities of airfoil geometries, we affirm that the database serves as a sufficiently comprehensive surrogate).

With the UIUC database taken as the global target set of airfoil shapes to be reconstructed by the DbM framework, we frame the identification of optimal baseline shapes as the problem of finding a subset of airfoil shapes whose morphing combinations can best approximate the entire UIUC database, thereby maximizing design-space coverage while minimizing design-space dimensionality. Mathematically, we let the set of available airfoil shapes in the UIUC database, after consistently discretizing them in the Selig-format vector form, be denoted by $\mathbb{A} \equiv \left\{ \vec{\mathcal{A}}_{\text{DB},1},~ \vec{\mathcal{A}}_{\text{DB},2},~ \cdots,~ \vec{\mathcal{A}}_{\text{DB},m} \right\}$, where $m$ (or $\#\mathbb{A}$) is 1,644 as of the present collection. We seek to select a subset of $n$ baselines, denoted as $\mathbb{B} \equiv \left\{ \vec{\mathcal{B}}_{1},~ \vec{\mathcal{B}}_{2},~ \cdots,~ \vec{\mathcal{B}}_{n} \right\} \subset \mathbb{A}$, such that the reconstruction capability measure $S^{\ddagger}$ of $\mathbb{B}$ over $\mathbb{A}$ is minimized.

The corresponding optimization problem can be generally expressed as:
\begin{equation}
\begin{aligned}
    \min_{\mathbb{B} \; \subseteq \; \mathbb{A}} \quad  S^{\ddagger} \quad \textrm{subject to} \quad  \#\mathbb{B}=n ,
    \label{eq:optimalbaseselectionproblem}
\end{aligned}
\end{equation}
where $n$ is the number of baseline airfoil shapes allowed in $\mathbb{B}$. When $n$ equals to $m$, the problem takes a trivial and global solution, $\mathbb{B} = \mathbb{A}$ (which evidently yields $S^{\ddagger} = 0$). Preferentially, $n \ll m$ to promote significant dimensionality reduction.

In Equation \ref{eq:optimalbaseselectionproblem}, $n$ (or $\# \mathbb{B}$) acts as a control parameter balancing the complexity and expressiveness of the design space. A larger $n$ increases the representational power but also the design-space dimensionality, whereas a smaller $n$ reduces the dimensionality at the cost of the airfoil reconstruction capacity. Therefore, solving Equation \ref{eq:optimalbaseselectionproblem} additionally aims to identify the smallest possible $n$ (or equivalently, the most compact set of baselines) that still achieves an acceptable level of reconstruction performance over the database. However, determining the acceptability is mostly done \textit{a posteriori}; thus, in the subsequent discussion, we presume that $n$ is given, for example, $n=10$.

\subsection{Approaches for the subset selection}

Using the concept of feature selection \parencite[][]{Guyon2003} or, similarly, factor screening \parencite[][]{Serani2024}, we identify the most influential elements among a large set of airfoil geometries. In other words, each individual airfoil shape in $\mathbb{A}$ is treated as a distinct feature that contributes to the construction of the overall design space. Since not all airfoil shapes are essentially unique --- some may offer redundant contributions to the representational capability --- an effective selection process should aim to retain only the most informative baselines in $\mathbb{B}$ while safely eliminating superfluous ones.

\subsubsection{Exhaustive search}
The most straightforward approach to baseline subset selection is to exhaustively compare all subsets. All possible combinations of $n$ baselines are enumerated from $\mathbb{A}$, and each candidate subset is evaluated based on its reconstruction capability measure $S^{\ddagger}$. The subset that minimizes $S^{\ddagger}$ is chosen as the optimal set.

For the UIUC database containing $m = 1,644$ airfoil shapes, the number of possible subsets of $\mathbb{A}$ for even modest values of $n$ becomes astronomical. For instance, selecting $n=10$ baselines would require evaluating approximately ${{1644}\choose{10}} \approx 3.9 \times 10^{25}$
candidate subsets. As each candidate subset's evaluation towards $S^{\ddagger}$ even necessitates solving Equation \ref{eq:reconstructionstandardopt} $m = 1,644$ times, such a number is computationally infeasible to process.

While exhaustive search guarantees identification of the globally optimal baseline sets for a given $n$, the combinatorial explosion in the number of subsets renders this approach unconditionally impractical for any realistic subset size. Thus, alternative strategies that significantly reduce the computational burden must be sought.

\subsubsection{Backward search}
One alternative is the backward search strategy. This method starts with the entire database as the initial baseline set, i.e., $\mathbb{B} = \mathbb{A}$. At each iteration, a single baseline is eliminated from $\mathbb{B}$ based on its relative contribution to the overall reconstruction capability, thereby reducing the size of the baseline set by one sequentially.

At the first iteration, all $m$ subsets of size $(m-1)$ are considered, where solving Equation \ref{eq:optimalbaseselectionproblem} requires only a single subordinate optimization in Equation \ref{eq:reconstructionstandardopt} for the eliminated airfoil of each subset. For each candidate subset, we evaluate $S^{\ddagger}$, and the subset that yields the smallest increase in $S^{\ddagger}$ from its initial value (zero, when $\mathbb{B} = \mathbb{A}$) is chosen. Then, in subsequent iterations, the elimination decision is guided differently to reduce computational cost. After solving Equation \ref{eq:reconstructionstandardopt} for each target airfoil based on the current baseline set, we compute the non-trivial morphing weight factors assigned to each baseline. The baseline whose total contribution, summed across all reconstructions, is then eliminated. In other words, we sequentially remove the baseline that contributes least to reconstructing the target airfoil set according to the absolute sum of its weight factors. This process is repeated iteratively until the number of remaining baselines reaches the desired subset size $n$.

The total number of times the subordinate optimization for airfoil reconstruction Equation \ref{eq:reconstructionstandardopt}, must be conducted throughout the process is given by \[m + \sum_{k=2}^{m-n+1}k = \frac{1}{2} \left\{m^2 +m(5-2n) + (n-3)n \right\},\] when counting only non-trivial cases (where a target airfoil is not included in the current baseline set). For $m = 1,644$ and $n = 10$, this amounts to approximately $1.3 \times 10^6$ optimal AirDbM weight factor searches. While this backward search strategy is inherently a greedy algorithm for solving Equation \ref{eq:optimalbaseselectionproblem} (i.e., making local optimal choices at each elimination), it offers a computationally tractable compromise between reconstruction accuracy and cost.

Despite substantially reducing the number of evaluations compared to exhaustive search, this strategy still incurs a computational cost that nearly scales $\mathcal{O}(m^2)$. As $m$ increases, the total number of subordinate optimizations becomes quadratically prohibitive. In the present case with $m=1,644$, the total number of evaluations remains an impractical computational burden.

\subsubsection{Forward search}\label{baselineselsub3}
Another approach is the forward search strategy, which we ultimately adopt in this study. Unlike backward search, forward search progressively builds the baseline set by sequentially adding airfoil shapes from the full database. The process starts with an empty set and, at each iteration, adds a single baseline that is expected to contribute most to improving the overall reconstruction of $\mathbb{A}$.

At the first iteration, all $m$ subsets of size 1, or equivalently, all $m$ individual airfoils are considered. For each, solving Equation \ref{eq:optimalbaseselectionproblem} simply requires summing the $(m-1)$ airfoil shape similarity measures with respect to all other $(m-1)$ airfoils, without any optimization, as no morphing needs to occur. The airfoil shape that yields the smallest $S^{\ddagger}$ is selected as the first baseline element. Then, in subsequent iterations, an airfoil shape that is least well reconstructed via the current baseline set (i.e., the one with the largest $S'$ value obtained from Equation \ref{eq:reconstructionstandardopt}) is added, until the desired subset size $n$ is reached.

Now, the total number of times Equation \ref{eq:reconstructionstandardopt} must be non-trivially solved is given by \[ \sum_{k=2}^{n-1} (m-k) = \frac{1}{2}(n-2)(2m-n-1), \] as the first iteration with a single baseline shape does not involve DbM weight optimization. For $m=1,644$ and $n=10$, this results in 13,108 evaluations, which finally becomes a computationally tractable number. Moreover, each reconstruction is performed with a low-dimensional input space because $k < n \ll m$, making the overall process significantly faster than the backward search.

In addition to its computational tractability --- since the number of evaluations scales linearly with the total database size $m$, i.e., $\mathcal{O}(m)$, given $m \gg n$ --- forward search offers a practical advantage for baseline selection. By sequentially adding baselines while observing the progressive improvement in reconstruction performance, it is possible to monitor how the design-space coverage evolves, allowing for us to flexibly control over how the final subset size $n$ be based on the observed performance trends during the search process, which is in line with the aim described in \S \ref{baselineselsub1}.

\subsection{Selected baselines}\label{sec:selbase}
To solve the optimization problems involving different baseline sets and target airfoils, as defined in Equation \ref{eq:reconstructionstandardopt}, a genetic algorithm (GA) was employed using the \texttt{Pymoo} framework \parencite{pymoo} with \texttt{Dask}-based parallelization \parencite{matthew_rocklin-proc-scipy-2015}. Each optimization run used a population size of 100 for up to 500 generations, with crossover and mutation operators set to simulated binary crossover and polynomial mutation, respectively, as provided by default in the framework. Parallel evaluation across multiple Dask workers (32 in this study) significantly accelerated the optimization process. Termination was based on convergence criteria evaluated over a rolling window of 20 generations: variable-space change (i.e., $\lVert\Delta \vec{w}_{\text{opt}}\rVert_\infty$ between successive generations less than $10^{-6}$) and objective-space change (i.e., $\Delta S'$ less than $10^{-8}$). We warm-started each optimization by including the previously obtained optimal weight vector (augmented with a zero morphing weight factor for the newly added baseline) as one of the initial population members when solving for the expanded baseline set.

\begin{table}[tb!]
\small
  \caption{Design-by-Morphing baseline airfoil shape set of size 12}
  \label{tab:baseline_airfoils}
  \begin{tblr}{
      colspec = {c X[c] l},
      rows={m, abovesep=8pt, belowsep=4pt},
      row{even} = {lightgray!10},
      row{1} = {rowsep = 2pt},
      width = \linewidth,
    }
    \hline \hline
    \textbf{Index} & \centering\textbf{Geometry} & \makecell[l]{\textbf{Airfoil Name} \\ \textbf{in the UIUC Database} \parencite{SeligUIUC}} \\
    \hline \hline
    $\mathcal{B}_1$ & \raisebox{-.5\height}{\centering\includegraphics[width=1.2in]{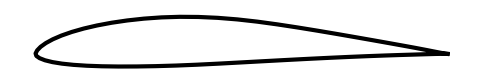}} & Eppler E195 \\
    $\mathcal{B}_2$ & \raisebox{-.5\height}{\centering\includegraphics[width=1.2in]{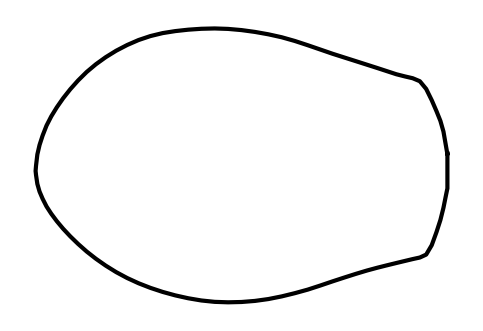}} & Wortman FX 79-W-660A \\
    $\mathcal{B}_3$ & \raisebox{-.5\height}{\centering\includegraphics[width=1.2in]{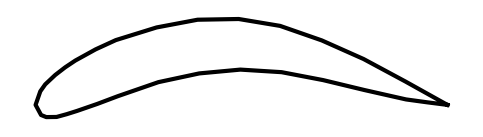}} & Gottingen 531 \\
    $\mathcal{B}_4$ & \raisebox{-.5\height}{\centering\includegraphics[width=1.2in]{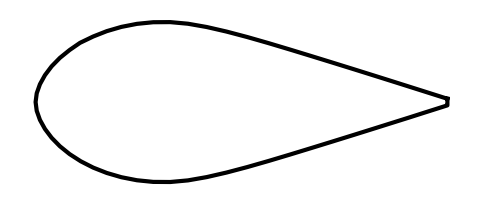}} & Eppler 864 Strut \\
    $\mathcal{B}_5$ & \raisebox{-.5\height}{\centering\includegraphics[width=1.2in]{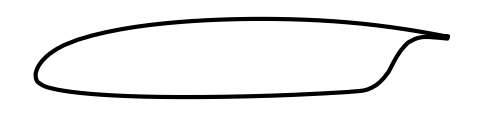}} & Roncz R1145MSM VariEze Canard Main \\
    $\mathcal{B}_6$ & \raisebox{-.5\height}{\centering\includegraphics[width=1.2in]{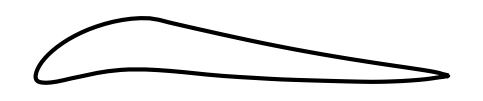}} & UIUC Chen \\
    $\mathcal{B}_7$ & \raisebox{-.5\height}{\centering\includegraphics[width=1.2in]{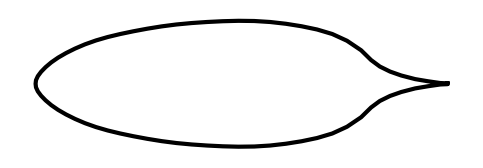}} & Griffith 30\% Suction \\
    $\mathcal{B}_8$ & \raisebox{-.5\height}{\centering\includegraphics[width=1.2in]{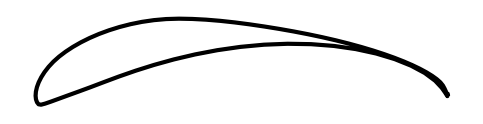}} & Selig S9104 \\
    $\mathcal{B}_9$ & \raisebox{-.5\height}{\centering\includegraphics[width=1.2in]{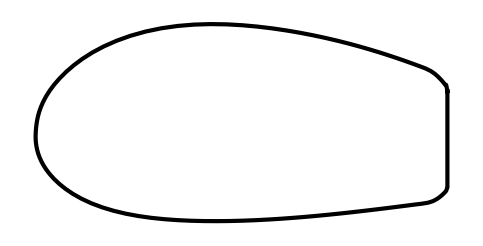}} & Althaus AH 93-W-480B \\
    $\mathcal{B}_{10}$ & \raisebox{-.5\height}{\centering\includegraphics[width=1.2in]{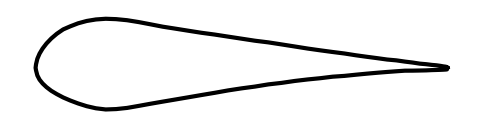}} & Althaus AH 81-K-144 W-F KLAPPE \\
    $\mathcal{B}_{11}$ & \raisebox{-.5\height}{\centering\includegraphics[width=1.2in]{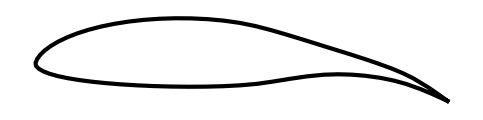}} & Eppler E664 (Extended) \\
    $\mathcal{B}_{12}$ & \raisebox{-.5\height}{\centering\includegraphics[width=1.2in]{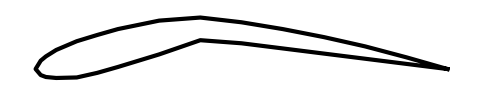}} & Saratov R/C Sailplane\\
    \hline \hline
  \end{tblr}
\end{table}

\begin{figure}[tb!]
    \centering
    \includegraphics[width=.6\linewidth]{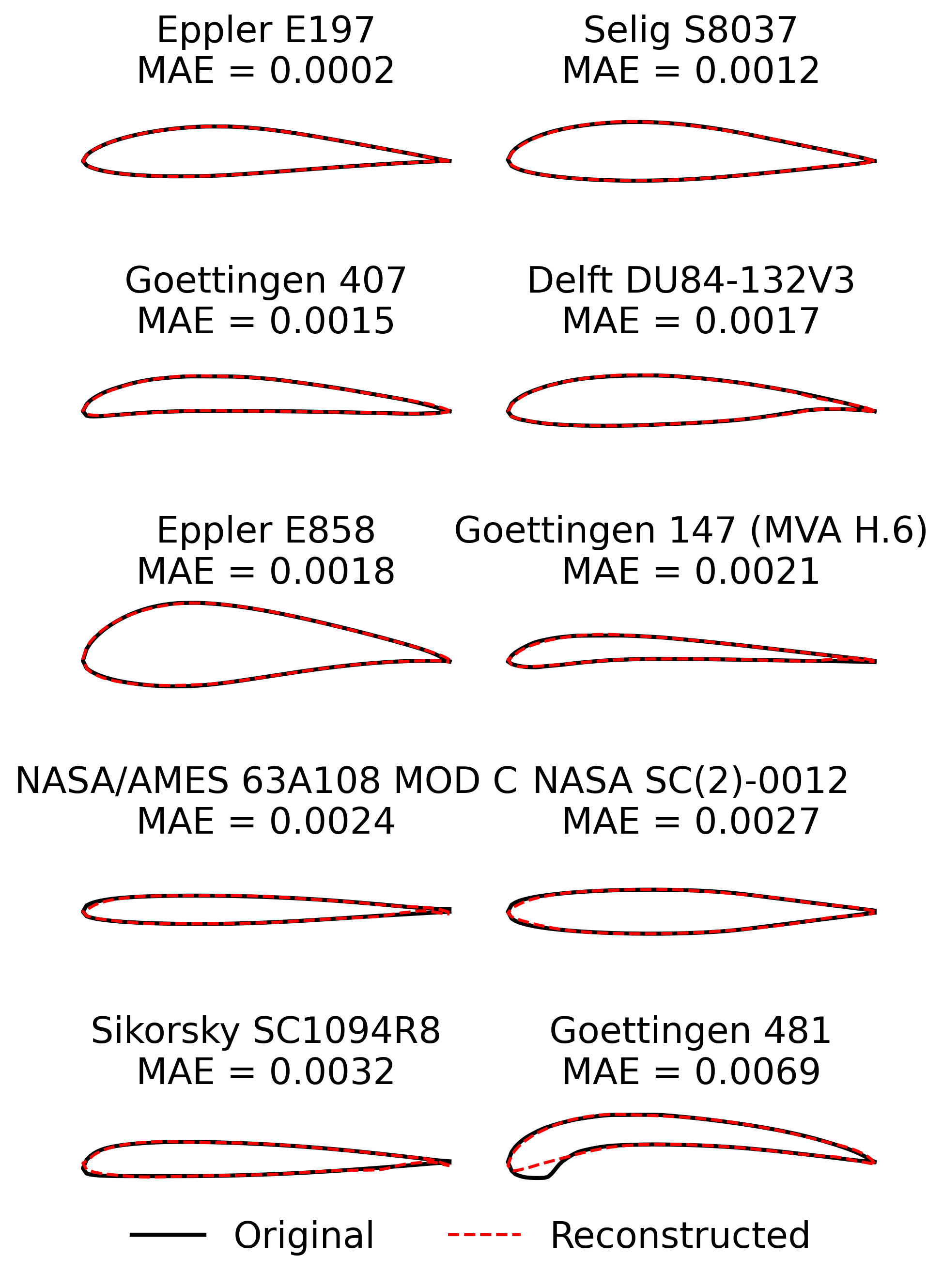}
    \caption{Comparison of original (black solid line) and reconstructed (red dashed line) airfoil geometries via Design-by-Morphing using the selected 12 baselines (see Table \ref{tab:baseline_airfoils}). The 10 airfoils displayed here are selected at equal rank intervals from best to worst reconstruction based on the Mean Absolute Error (MAE) similarity metric (see Equation \ref{eq:similarity_5}).}
    \label{fig:AirfoilReconstruct}
\end{figure}

Throughout the forward search strategy that is powered by GA, the optimal baseline set of size $n=12$ was identified, which are presented in Table \ref{tab:baseline_airfoils}. The index indicates the order in which each airfoil was added to the baseline set during the forward search. Accordingly, an optimal baseline set of any smaller size $\eta < 12$ can be constructed by considering only the first $\eta$ airfoils from this table (that is, baselines \#1 -- \#$\eta$). With these 12 baseline shapes, the DbM approach successfully reconstructed all 1,644 airfoil shapes in the database with $S'$ (hereafter used interchangeably with mean absolute error, or MAE) below 0.01. Figure \ref{fig:AirfoilReconstruct} illustrates the comparison between the original and DbM-reconstructed airfoil geometries for 10 airfoils selected at equal rank intervals from the best to worst MAE. The best-performing reconstruction, observed for Eppler E197, exemplifies the inherent redundancy within the database. The geometry is nearly similar to the first baseline shape, Eppler E195, albeit with a slight variation in camber thickness, demonstrating that not all airfoil shapes represents truly unique design features. This underscores the motivation for the current practice of dimensionality reduction.

On the other hand, in Figure \ref{fig:AirfoilReconstruct}, the worst reconstruction case of Gottingen 481 reveals the limited coverage imposed by the reduced baseline set. There exists non-negligible deviation in shape, particularly due to its pronounced curvature at the bottom surface around the leading edge, which suggests that such a geometric feature is not fully taken into account by the current 12 baselines. In compliance with the forward search strategy, we could additionally consider Gottingen 481 as a potential additional baseline shape to better encompass such \textit{significantly bent} airfoil shapes within the design space, while increasing the design-space dimensionality by one. The decision to augment the baseline set with such shapes should require a careful consideration of the balance between design space diversity and the minimal design-space dimensionality. This trade-off can be informed by analyzing the trend of reconstruction convergence as the number of baseline shapes increases.

\begin{figure}
    \centering
    \includegraphics[width=0.8\linewidth]{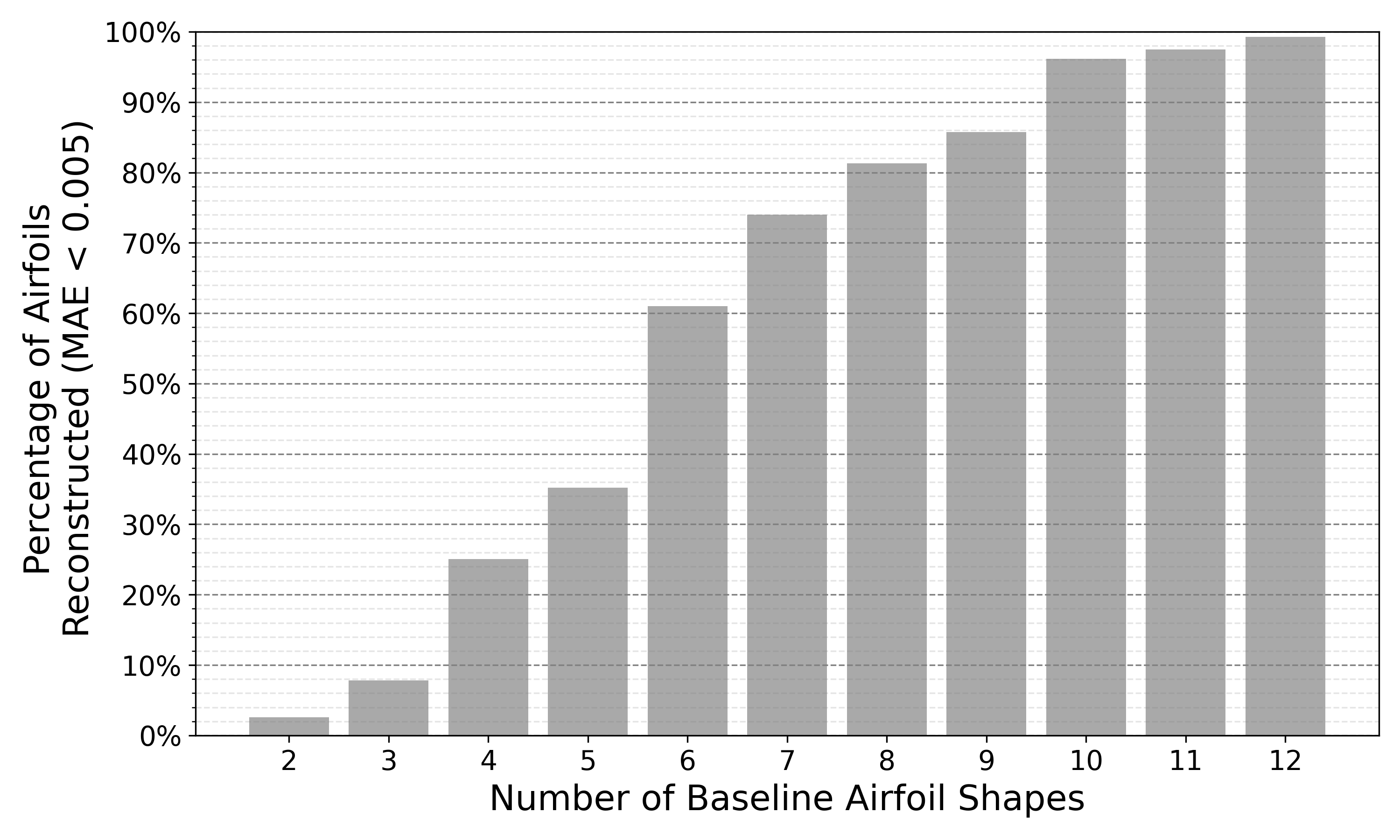}
    \caption{Percentage of airfoils in the database that can be reconstructed via Design-by-Morphing with an Mean Absolute Error (MAE) below 0.005 for baseline set sizes from 2 to 12 (see Table \ref{tab:baseline_airfoils}).}
    \label{fig:ConvergenceTrend}
\end{figure}

The effectiveness of the selected 12 baseline shapes is supported by the convergence trend of the reconstruction rate, as visualized in Figure \ref{fig:ConvergenceTrend}. This plot illustrates the percentage of airfoil shapes in the database reconstructed within a MAE tolerance of 0.005 as the number of baseline shapes varies from 2 (i.e., baselines \#1 -- \#2) to 12 (i.e., baselines \#1 -- \#12). This tolerance threshold is chosen based on its previous use in \textcite{Sheikh2023} and, as our previous comparison plots demonstrate, it approximately marks the level at which visually notable discrepancies between original and reconstructed shapes become apparent (compare the second-worst case (MAE = 0.0032) to the worst case (MAE = 0.0069) in Figure \ref{fig:AirfoilReconstruct}). As the number of baselines increases, the reconstruction rate increases with a flattening of the curve beyond 10 baselines, which indicates diminishing returns for further increases in dimensionality. 

Moreover, with the selected set of 12 baselines, more than 98 \% of airfoil shapes were reconstructed using DbM within an MAE tolerance of 0.005. This level of reconstruction matches the performance reported in our previous DbM study on airfoil optimization \parencite{Sheikh2023}, which relied on 25 baseline shapes. Thus, the current selection successfully achieves comparable reconstruction quality while reducing the design-space dimensionality by more than half, from 25 to 12.

{\color{black} Given that the selected 12-baseline set almost entirely spans the UIUC database design space, one might expect that randomly selecting 12 baselines could maintain reconstruction capacity while bypassing the search cost, analogous to a change of basis in linear vector spaces. However, DbM's morphing process is inherently nonlinear, so replacing shapes would not guarantee preservation of space-spanning capability. To verify this, we tested three randomly selected 12-baseline sets. Their successful reconstruction rates (MAE < 0.005) were significantly lower and inconsistent, achieving only 75\%, 66\%, and 49\%, respectively. These outcomes confirm that a systematic search is essential to achieve both high performance and consistency. In any case, users can bypass this search process entirely and directly utilize the validated baseline set provided in Table \ref{tab:baseline_airfoils}.}

{\color{black} Lastly, it is worth recalling that the current baseline set selection process assumes the UIUC database to be globally representative. However, one can point out its limited coverage of supersonic or hypersonic airfoils. Similar to the Gottingen 481 case, the DbM framework can readily accommodate baseline augmentation when applications require non-inclusive or underrepresented designs, such as diamond-shaped supersonic airfoils \parencite{Jernell1974}, which seamlessly expands the global design space. This inherent adaptability demonstrates the framework's strength in accommodating unforeseen or novel designs.}

\subsection{Design capacity comparison}
While design capacity comparisons of DbM against conventional airfoil parameterization methods constituted the main theme of our previous work \parencite{Sheikh2023}, which confirmed DbM's competitiveness with methods specifically designed for airfoils, we provide a brief comparison here again to evaluate the performance of the present reduced 12-baseline DbM (henceforth denoted AirDbM to specify its application to airfoil design and optimization and its reduced baseline set of 12 as defined in Table \ref{tab:baseline_airfoils}). 

To assess AirDbM's design capacity under reduced dimensionality constraints, we compared reconstruction performance using a consistent number of design variables across all methods: 12 design variables for AirDbM, Hicks–Henne bump functions, class-shape transformation (CST), and the parametric section (PARSEC) method, with 13 variables for non-uniform rational B-splines (NURBS) due to its formulation requirements. Implementation details are provided in Appendix \ref{app:AP}. 
The comparison was conducted across all 1,644 airfoils in the UIUC database using the same MAE evaluation scheme as established in the earlier sections.

\begin{table}[tb!]
    \centering
    \caption{\color{black}Design capacity comparison across airfoil design methods.}
    \label{tab:design_capacity}
    \begin{tblr}{
      colspec = {c c c},
      rows={m, abovesep=8pt, belowsep=4pt},
      row{even} = {lightgray!10},
      row{1} = {rowsep = 2pt},
      width = \linewidth
    }
    \hline \hline
    \textbf{Method} & \makecell[c]{\textbf{Overall MAE} \\ \textbf{(Mean ± Std.)}} & \makecell[c]{\textbf{Percentage of Airfoils} \\ \textbf{Reconstructed (MAE < 0.005)}} \\
    \hline \hline
    CST & 0.0012 ± 0.0008 & 99.4 \% \\
    \textbf{AirDbM} & \textbf{0.0021 ± 0.0009} & \textbf{99.3 \%} \\
    NURBS & 0.0024 ± 0.0019 & 91.6 \% \\
    PARSEC & 0.0032 ± 0.0025 & 84.5 \% \\
    Hicks-Henne & 0.0053 ± 0.0035 & 58.7 \% \\
    \hline \hline 
\end{tblr}
\end{table}

Table \ref{tab:design_capacity} presents the quantitative comparison results. AirDbM shows competitive performance, achieving reconstruction quality comparable to CST while significantly outperforming NURBS, PARSEC, and Hicks-Henne. This performance is particularly significant given that conventional methods like NURBS and Hicks-Henne experience substantial degradation in design coverage when constrained to lower dimensionalities, compared to the higher-dimensional cases (24-26 design variables) achieving 98 \% reconstruction success reported in \textcite{Sheikh2023}. The superior performance of CST appears to come from its aerodynamic-specific design principles, employing carefully crafted class and shape functions with mathematical rigor tailored for airfoil applications \parencite[see][]{Kulfan2006}. AirDbM's comparable performance is therefore noteworthy, considering its universal morphing principle that remains applicable across diverse design domains while maintaining high reconstruction fidelity under significant dimensionality reduction.

\begin{figure}[tb!]
    \centering
    \includegraphics[width=\linewidth]{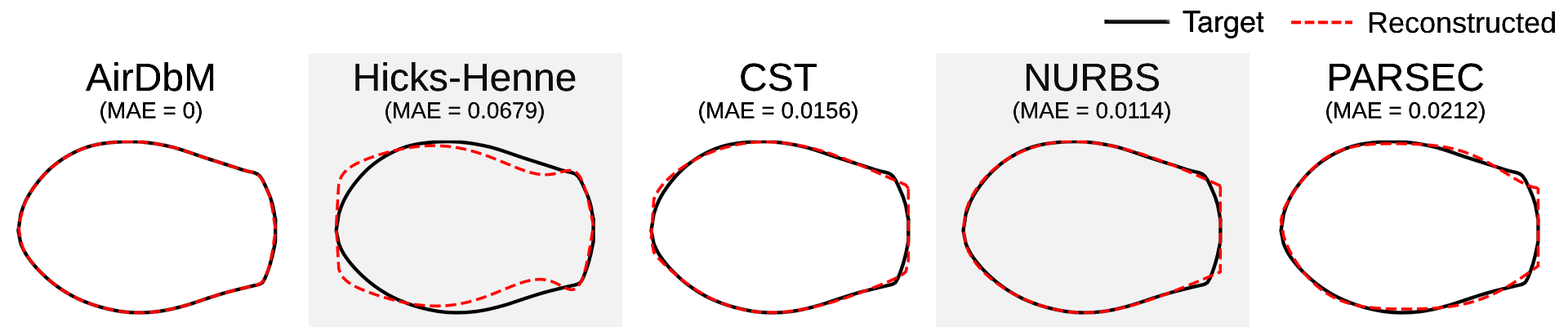}
    \caption{{\color{black} Reconstruction of Wortman FX 79-W-660A airfoil. Depicted are target (black solid line) versus reconstructed (red dashed line) airfoils for the present 12-baseline Design-by-Morphing (AirDbM), Hicks-Henne, class-shape transformation (CST), non-uniform rational B-spline (NURBS), and parametric section (PARSEC) methods.}}
    \label{fig:reconstruct_comparison}
\end{figure}

To illustrate the practical implications of the design capacity differences, Figure \ref{fig:reconstruct_comparison} presents reconstruction results for the Wortman FX 79-W-660A airfoil, which represents the worst-reconstructed case (i.e., highest summed MAE across all five methods). This result is primarily due to its unusually thick profile which deviates from typical airfoil geometries. Although conventional methods struggle with this unconventional shape, AirDbM naturally overcomes this challenge by directly incorporating the specific thick profile as its baseline ($\mathcal{B}_2$). The key strength of the DbM framework lies in the ease of incorporating such novel designs contributing to design diversity: regardless of how unconventional a target shape may be, it can be effectively utilized in the design process with the cost of only a single design parameter. Readers are encouraged to refer to a similar discussion in our previous work regarding the reconstruction of a `mirrored' airfoil with flipped sharp and blunt edges \parencite[][pp. 1447-1448]{Sheikh2023}.

\section{Performance Evaluation}\label{airfoilshapeopt}
\subsection{Multi-objective airfoil optimization}\label{airmultiopt}
We first incorporate AirDbM into an airfoil shape optimization problem, optimizing the airfoil shape based on aerodynamic information obtained from a flow solver. {\color{black} Since the primary objective of this test study is to evaluate the computational efficiency of the proposed DbM approach with reduced design-space dimensionality, the focus is placed more on analyzing optimization performance than on the optimal outcomes themselves. Accordingly, we revisit the airfoil optimization setup from our previous work \parencite{Sheikh2023} to investigate the optimization performance under reduced design-space dimensionality, compared with the previous 25-baseline case}. It should be noted that multi-objective optimization for airfoil dynamics has garnered increasing attention in recent years \parencite[e.g.,][]{Jing2023,Zhang2024,Jung2024}, further underscoring the practical utility of this work.

\begin{figure}
    \centering
    \includegraphics[width=0.95\linewidth]{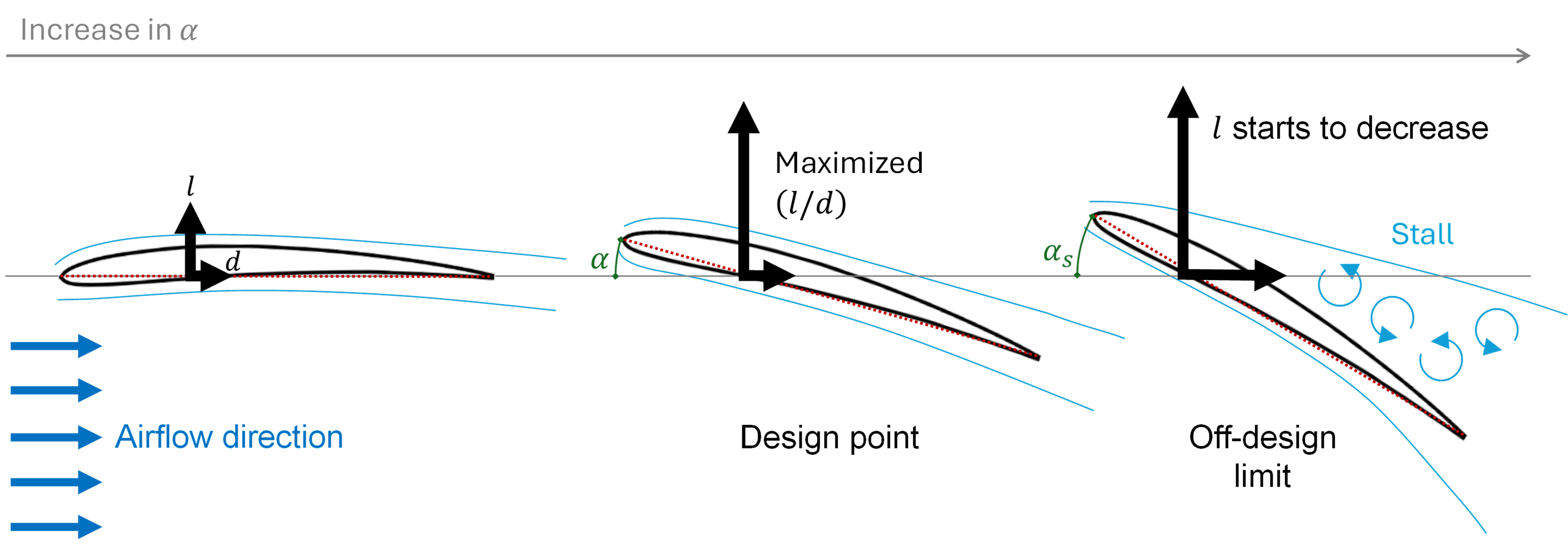}
    \caption{Airfoil performance with increasing angle of attack $\alpha$, depicting changes in lift $l$ and drag $d$. The figure illustrates the design point for maximum lift-to-drag ratio $(l/d)_{\max}$ and the stall tolerance, quantified by the angle of attack range from the design point to the off-design limit of stall at $\alpha_s$.}
    \label{fig:AirMultiOptScheme}
\end{figure}

An airfoil with chord length $c$, subjected to a freestream flow of speed $U$, fluid density $\rho$, and kinematic viscosity $\nu$ (i.e., under the Reynolds number condition $\text{Re} = Uc/\nu$), is characterized by two key dynamic performance parameters: the lift coefficient $C_l$ and the drag coefficient $C_d$, defined as:
\begin{equation}
    C_l (\alpha) = \frac{2l(\alpha)}{\rho U^2 c},
\end{equation}
\begin{equation}
    C_d (\alpha) = \frac{2d(\alpha)}{\rho U^2 c},
\end{equation}
where $l$ and $d$ represent the lift and drag forces per unit span, respectively, both being functions of the airfoil's angle of attack $\alpha$. The lift coefficient $C_l (\alpha)$ typically increases with $\alpha$ at low angles until stall occurs, beyond which it decreases. The stall angle $\alpha_s$ thus can be defined as the first local maximum of $C_l$ while increasing $\alpha$ from $0^{\circ}$. Following \textcite[][]{Sheikh2023}, we then consider the optimization of the following two composite objectives based on $C_l$ and $C_d$:
\begin{equation}
    (l/d)_{\max} \equiv\max_{\alpha} \frac{ C_l (\alpha )} { C_d (\alpha)},
\end{equation}
\begin{equation}
    \Delta \alpha \equiv \max \left( 0, ~ \alpha _s - \argmax_{\alpha} \frac{ C_l (\alpha )} { C_d (\alpha)} \right),
\end{equation}
where first objective $(l/d)_{\max}$ is the maximum lift-to-drag ratio under the design operating condition (i.e., $\alpha$ associated with the maximum $l/d$), and the second $\Delta \alpha$ quantifies the stall tolerance, representing the angle of attack range for off-design operations. They are illustrated in Figure \ref{fig:AirMultiOptScheme}.

These aerodynamic objectives are evaluated using XFOIL 6.99, a widely-accepted inviscid/viscous zonal airfoil analysis program \parencite{Drela1989} for quick initial design studies, at $\text{Re} = 10^6$ {\color{black} under incompressible flow conditions suitable for subsonic flows with negligible air density variation ($\text{Ma} = U/c_s \ll 1$, where $c_s \approx 3 \times 10^2~\text{m/s}$ is the speed of sound).} The low computational cost of this solver enables direct exploration of the objective space (without the need for a surrogate model). Accordingly, we use the MATLAB-based non-dominated sorting genetic algorithm (NSGA-II) \texttt{gamultiobj} \parencite{996017}, by taking each morphed airfoil shape $\mathcal{M}$'s 12 DbM weight factors $(w_1,~\cdots,w_{12})$ as its genetic representation. The detailed setup of the optimization is provided in Appendix \ref{app:MO}. {\color{black} Readers are also recommended to refer to \textcite[][{see Appendices A and B}]{Sheikh2023}, which includes preliminary validation steps of the setup we replicate here for the sake of comparison.  For instance, to ensure evaluation robustness, our XFOIL implementation employs convergence check strategies such as restarting with fresh initial guesses and correctness verification through viscous-inviscid drag coefficient comparison. These metrics and evaluation procedures were validated against the reference XFOIL evaluation database \parencite{airfoiltools}.}

Before diving into the analysis of the results, it is pertinent to distinguish expected behaviors from unusual improvements when using the reduced baseline set. Generally, decreasing design variables should accelerate convergence, requiring fewer total generations. Our previous optimization with 25 baselines ran for 3,000 GA generations, and we anticipate the current one to take fewer. However, faster convergence does not guarantee a superior or even equivalent Pareto front. Dimensionality reduction inevitably compacts the design space, likely leading in our case to the minimal space encompassing the existing database. Consequently, the common expectation is that the resulting Pareto front will hardly outperform that from a larger baseline set (assuming sufficient convergence). If the optimization with the reduced design space yields enhanced solutions that dominate prior Pareto-optimal solutions, this would constitute a key improvement.

\begin{figure}[tb!]
    \centering
    \includegraphics[width=0.8\linewidth]{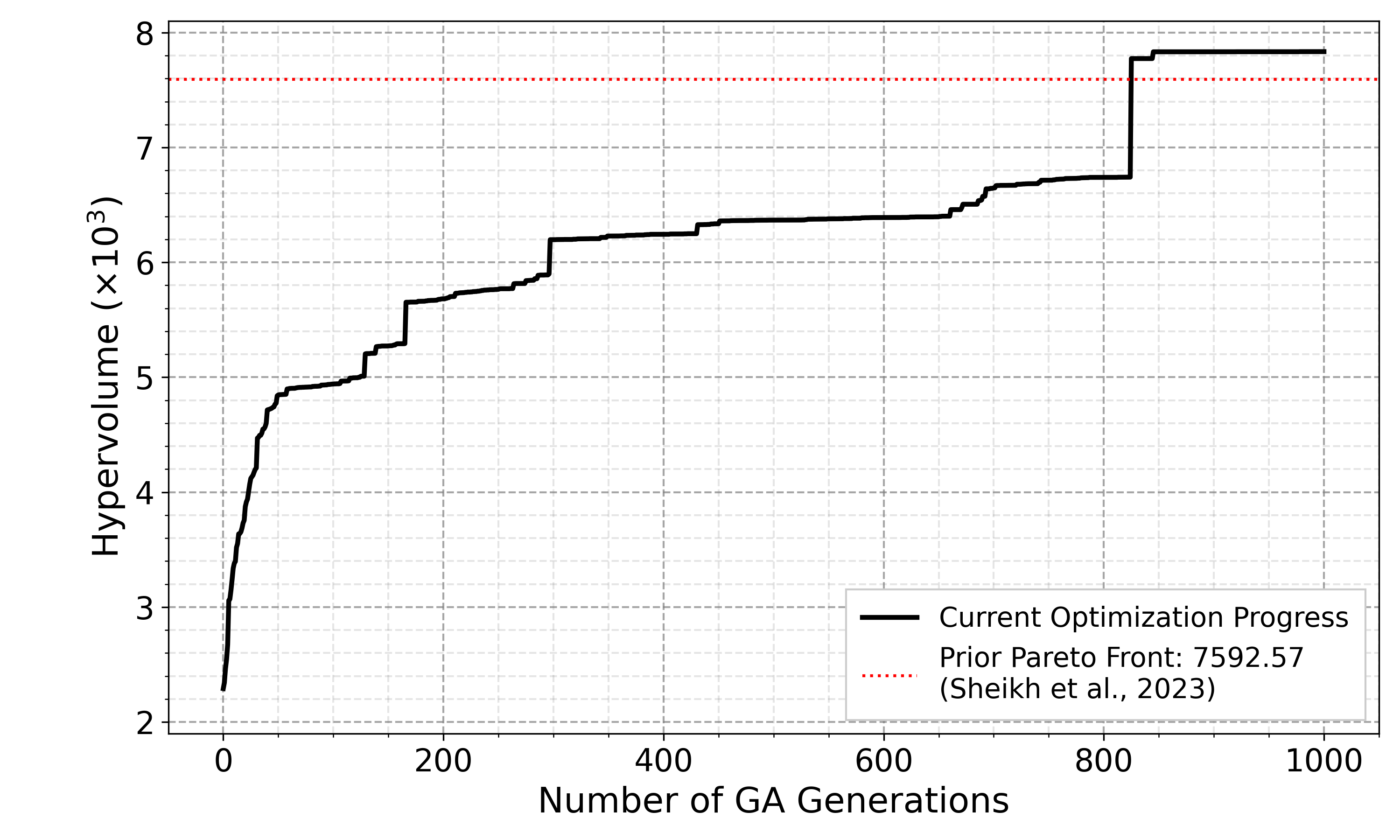}
    \caption{Hypervolume progression for the multi-objective airfoil optimization using AirDbM, tracking the hypervolume with respect to genetic algorithm (GA) generations (solid black line), demonstrating improved Pareto front quality. The dotted red line indicates the hypervolume from the prior study with 25 baselines \parencite[see][p. 1450]{Sheikh2023}.}
    \label{fig:HyperVolumeEvolution}
\end{figure}

Figure \ref{fig:HyperVolumeEvolution} illustrates the progression of the hypervolume indicator --- a widely adopted metric in multi-objective optimization for evaluating the quality of a set of non-dominated solutions (i.e., solutions for which no objective can be improved without degrading at least one other objective) \parencite[see][]{Li2020, Guerreiro2022} --- throughout the GA generations for the current optimization. In the present bi-objective context, the hypervolume of a set of non-dominated solutions $((l/d)_{\max,1}, \Delta \alpha_{1})$, $\cdots$, $((l/d)_{\max,k}, \Delta \alpha_{k})$ is defined as the total area of the following 2D Pareto dominance region $\mathcal{R}$:
\begin{equation}
    \mathcal{R} = \bigcup_{i=1,\,\cdots\,,\,k} \left\{ (f_1, f_2) \in \mathbb{R}^2 ~\Big|~ 0 \le f_1 \le (l/d)_{\max,i} ~\text{and}~ 0 \le f_2 \le \Delta \alpha_i \right\},
\end{equation}
where the origin $(0,\,0)$ is taken as the reference (nadir) point. The observed evolutionary trend of the hypervolume in Figure \ref{fig:HyperVolumeEvolution}, characterized by a steep initial increase in hypervolume followed by a more gradual convergence with sporadic leaps, is consistent with typical performance patterns reported in NSGA-II literature \parencite[e.g.,][]{Steuler2020, Antoniou2020}.

Notably, the hypervolume achieved by the current optimization with 12 baselines surpasses the final hypervolume of 7592.57 in \textcite{Sheikh2023}, which utilized 25 baseline airfoils, at approximately the 850th generation. This milestone is achieved significantly earlier than the 3000 generations run in the previous study, underscoring the expedited convergence attributable to the reduced design-space dimensionality. While the outperformance in terms of the hypervolume indicator is a positive indication of the efficacy of AirDbM in navigating the design space, it is important to note that this is merely one measure of Pareto front quality. Accordingly, a detailed comparison of the Pareto fronts, crucial for understanding the specific trade-offs associated with the current, more compact design space, should follow.

\begin{figure}[tb!]
    \centering
    \includegraphics[width=0.65\linewidth]{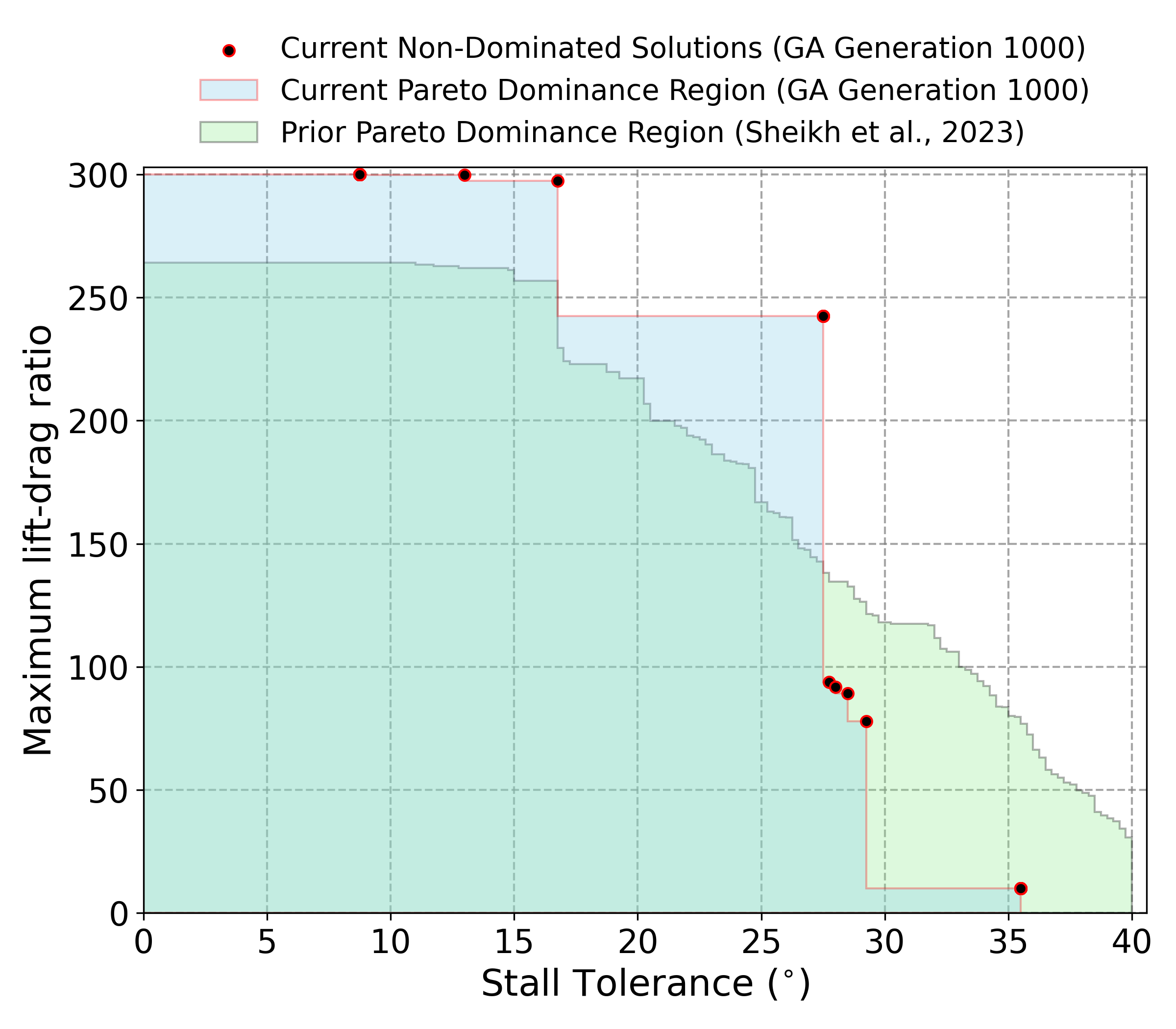}
    \caption{Pareto front comparison for airfoil optimziation: AirDbM (current) versus prior work \parencite[see][p. 1450]{Sheikh2023}.}
    \label{fig:ParetoFront}
\end{figure}

A direct comparison of the Pareto front obtained using the current AirDbM approach after the 1000 GA generations against that from \textcite{Sheikh2023} is presented in Figure \ref{fig:ParetoFront}. AirDbM successfully identifies new non-dominated solutions achieving significantly higher $(l/d)_{\max}$, particularly at moderate stall tolerances, thereby dominating the prior Pareto front in this portion of the objective space. Nonetheless, Figure \ref{fig:ParetoFront} also reveals that the current Pareto front does not extend to the same stall tolerance values achieved by the prior study, which found non-dominated solutions approaching $\Delta \alpha \approx 40^{\circ}$. This suggests that while the 12 selected baselines enable efficient design exploration and yield improvements in certain regions, they may not possess the geometric diversity required to reproduce solutions at the extreme end of the $\Delta \alpha$ spectrum previously accessible with 25 baselines.

The inability to reach these high $\Delta \alpha$ solutions is further evidenced by attempts to reconstruct specific Pareto-optimal airfoils from the prior study. For instance, when reconstructing the prior optimal airfoil solution characterized by the highest $\Delta \alpha$ using AirDbM, the reconstruction resulted in MAE exceeding the 0.005 threshold, indicating significant discrepancies. Considering that much finer geometric tolerances, on the order of $10^{-4}$, e.g., Kulfan's wind-tunnel tolerance \parencite{Kulfan2006, Masters2017}, are regarded as necessary to ensure the replication of aerodynamic performance, it is likely that the observed truncation in the current Pareto front for high stall tolerance implies the bounds imposed by the reduced geometric variability of the AirDbM design space. Nevertheless, given the substantial decrease in lift-to-drag ratios typical in 3D wing applications (usually by an order of magnitude), the enhanced Pareto front in $(l/d)_{\max}$ found by AirDbM can offer greater practical utility to offset 3D decrease.



\begin{figure}[tb!]
    \centering
    \includegraphics[width=\linewidth]{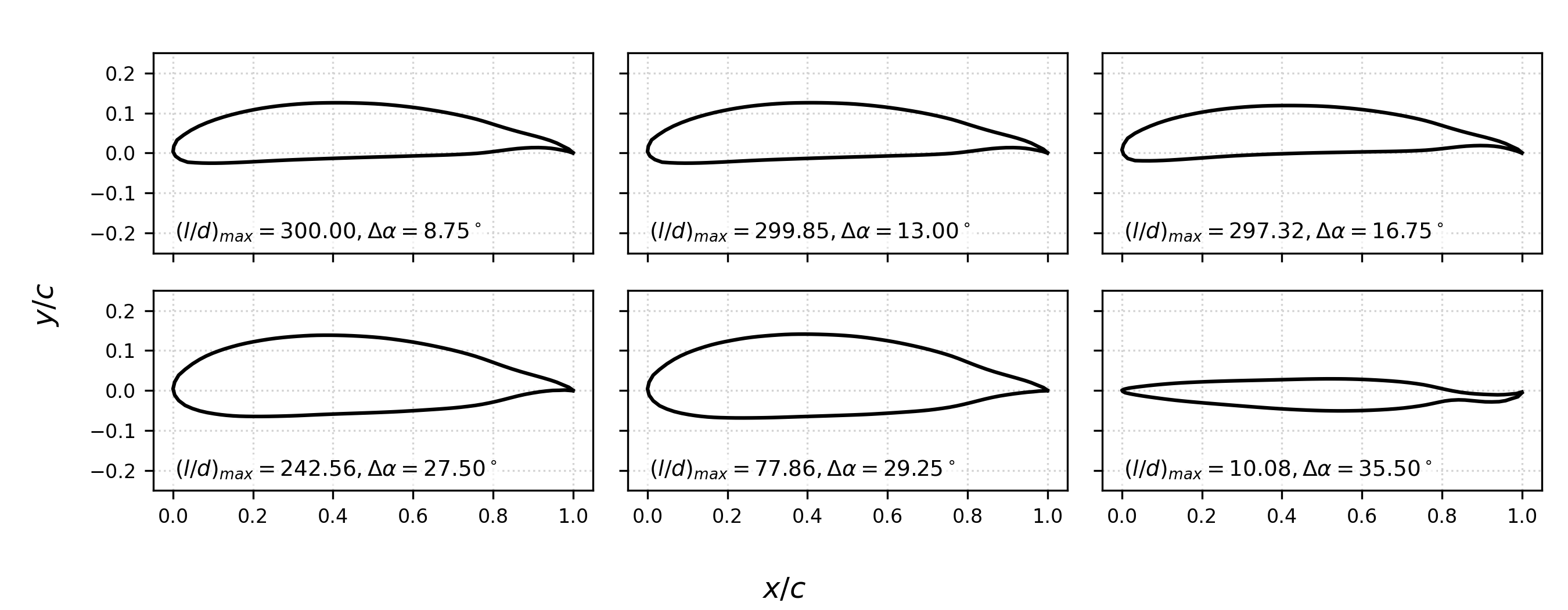}
    \caption{{\color{black} A selection of airfoil shapes from AirDbM's Pareto front displayed in Figure \ref{fig:ParetoFront}, arranged in descending order of $(l/d)_{\max}$ from top-left to bottom-right.}}
    \label{fig:ParetoOptimalAirfoils}
\end{figure}

{\color{black} 
Figure \ref{fig:ParetoOptimalAirfoils} showcases a selection of Pareto-optimal airfoil shapes obtained from the current AirDbM optimization. Excluding clustered solutions with minimal geometric differences from the presented one with $(l/d)_{\max} = 77.86$ and $\Delta \alpha = 29.25^\circ$, we present six representatives that capture the range of trade-offs between $(l/d)_{\max}$ and $\Delta \alpha$.

The first three airfoils (top row) exhibit similar thin-profile geometries, achieving high lift-to-drag ratios ($(l/d)_{\max} = 300.00$, $299.85$, and $297.32$) that represent improvements over our previous Pareto front. The $\Delta \alpha$ variance among these airfoils falls within the expected range for high $(l/d)_{\max}$ optimal airfoil groups identified in our previous work. The fourth airfoil ($(l/d)_{\max} = 242.56$ and $\Delta \alpha = 27.50^\circ$) demonstrates a thicker profile that achieves greater stall tolerance at the expense of lift-to-drag ratio. This thickness-induced performance trade-off is consistent with observations from \textcite{Sheikh2023}.

The final two airfoils, while non-dominant compared to our previous Pareto front, offer instructive insights into optimization behavior. The fifth airfoil shares a geometric appearance similar to the fourth, albeit slightly thicker, but exhibits a substantial performance drop ($(l/d)_{\max} = 77.86$), implying the highly nonlinear nature of aerodynamic performance. However, this can stem from potential limitations in XFOIL's solution accuracy --- even when converged, performances may represent physically irrelevant solutions arising from the simplified modeling inherent to preliminary design tools. The sixth airfoil presents a distinctly different morphology: a spear-like sharp profile ($(l/d)_{\max} = 10.08$, $\Delta \alpha = 35.50^\circ$). This solution presumably represents a physically unrealistic configuration that emerges from XFOIL's inherent modeling simplifications in preliminary aerodynamic analysis.
}

{\color{black} For readers interested in the physical aspects of optimal airfoil designs resulting from this optimization, it should be carefully taken into account that, due to XFOIL's 2D nature that cannot completely capture real-world 3D wing effects (e.g., tip effects, wakes, and spanwise separations), predicted $(l/d)_{\max}$ and $\Delta \alpha$ values can be excessively elevated compared to actual 3D applications. While the scope of the current test study is limited to validating design-space dimensionality effectiveness, future work will replace this preliminary-level solver with higher-fidelity solvers directly solving the Navier-Stokes equations, such as three-dimensional Reynolds-averaged Navier-Stokes (RANS) simulations. This will enable detailed physics-based analysis of solutions, building upon the framework robustness demonstrated in the current study.}

\subsection{Airfoil geometry learning}\label{airfoilgeomgen}
Through our practice of dimensionality reduction in \S \ref{baselinesel}, AirDbM has demonstrated a design capacity comparable to its predecessor using a larger baseline set and achieves a design span on par with or superior to several conventional airfoil parameterization methods with a consistent number of design variables. This similar database reconstruction rate with significantly fewer design variables underscores its effectiveness. 

Additionally, from a designer's perspective, adaptability is considered as crucial as effectiveness. In this example, we assume that adaptability across different methods could be assessed by observing designers, initially unfamiliar with airfoil parameterization, while they iteratively generate shapes and improve their learning in an empirical manner. However, using \textit{human} designers could present challenges in validating their level of unfamiliarity, leading to uncontrollable biases in the evaluation.

\begin{figure}[tb!]
    \centering
    \includegraphics[width=0.75\linewidth]{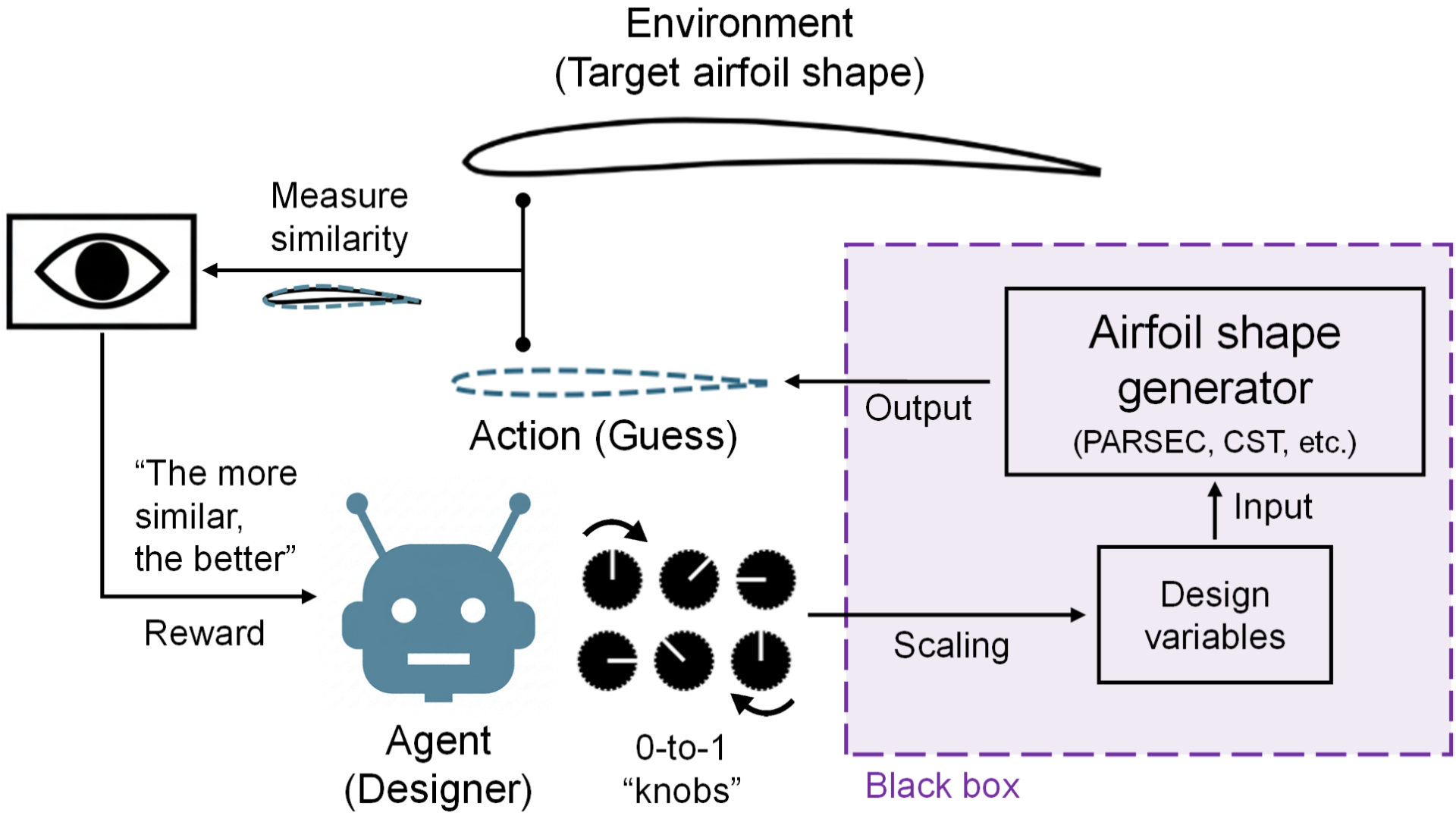}
    \caption{Reinforcement learning framework for airfoil geometry generation `game,' where the agent (designer) manipulates normalized control inputs (tuning `knobs') for a black-box airfoil shape generator. The agent acts to generate an airfoil `guess' without \textit{a priori} insight into the generator's internal process, and receives a reward based on similarity to the `target' airfoil shape provided in the environment.}
    \label{fig:AirfoilGenGameDiagram}
\end{figure}

Instead, a \textit{machine} agent driven by recent advancements in reinforcement learning (or neuro-dynamic programming) algorithms offers a compelling alternative to serve as an unbiased and initially `ignorant' designer. Reinforcement learning (RL) enables an agent to learn optimal behavior through trial-and-error interactions within an environment by maximizing a reward signal \parencite{Bertsekas2019}. It has seen increasing application in aerodynamic design optimization problems demanding intelligence and experience \parencite[e.g.,][]{Hui2021, Patel2024}. In this study, we utilize this framework for airfoil geometry generation, where the agent plays a `game' of guessing the control inputs for a \textit{black-box} airfoil shape generator to match the output airfoil shape with the given target airfoil shape. Over multiple iterations, the agent gets empirical knowledge about input-output relations, resulting in getting a more similar guess to the target. The overall framework scheme is illustrated in Figure \ref{fig:AirfoilGenGameDiagram}.

\begin{figure}[tb!]
    \centering
    \begin{subfigure}{\linewidth}
        \centering\renewcommand{\thesubfigure}{\Alph{subfigure}}
        \includegraphics[width=\linewidth]{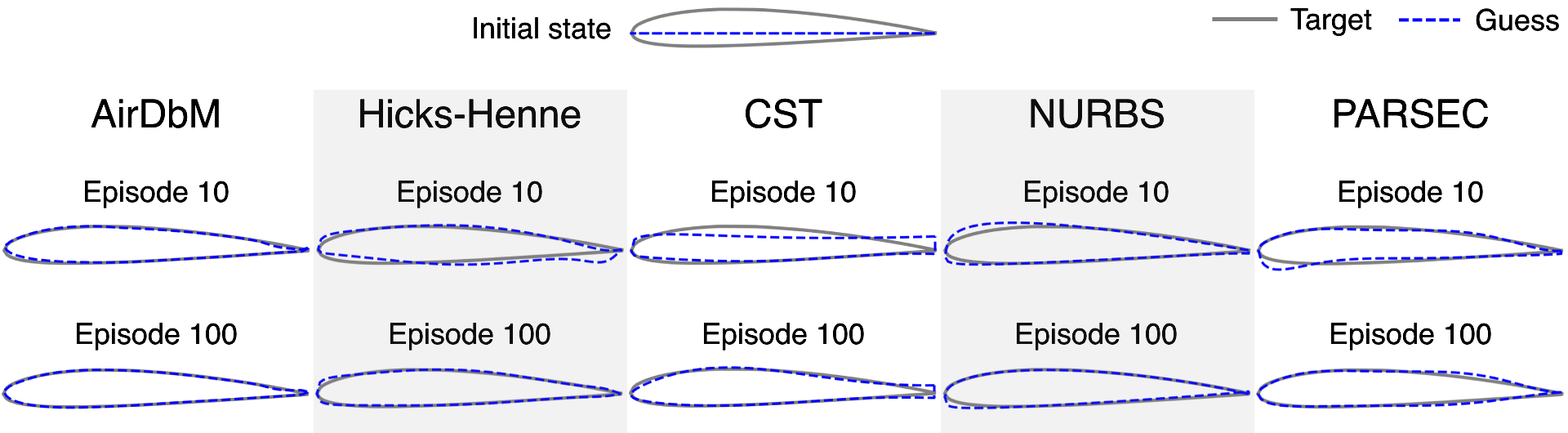}
        \captionsetup{justification=centering}\caption{NACA 2412}
        \label{fig:AirfoilGenGeom-a}
    \end{subfigure}
    \\[10pt]
    \begin{subfigure}{\linewidth}
        \centering\renewcommand{\thesubfigure}{\Alph{subfigure}}
        \includegraphics[width=\linewidth]{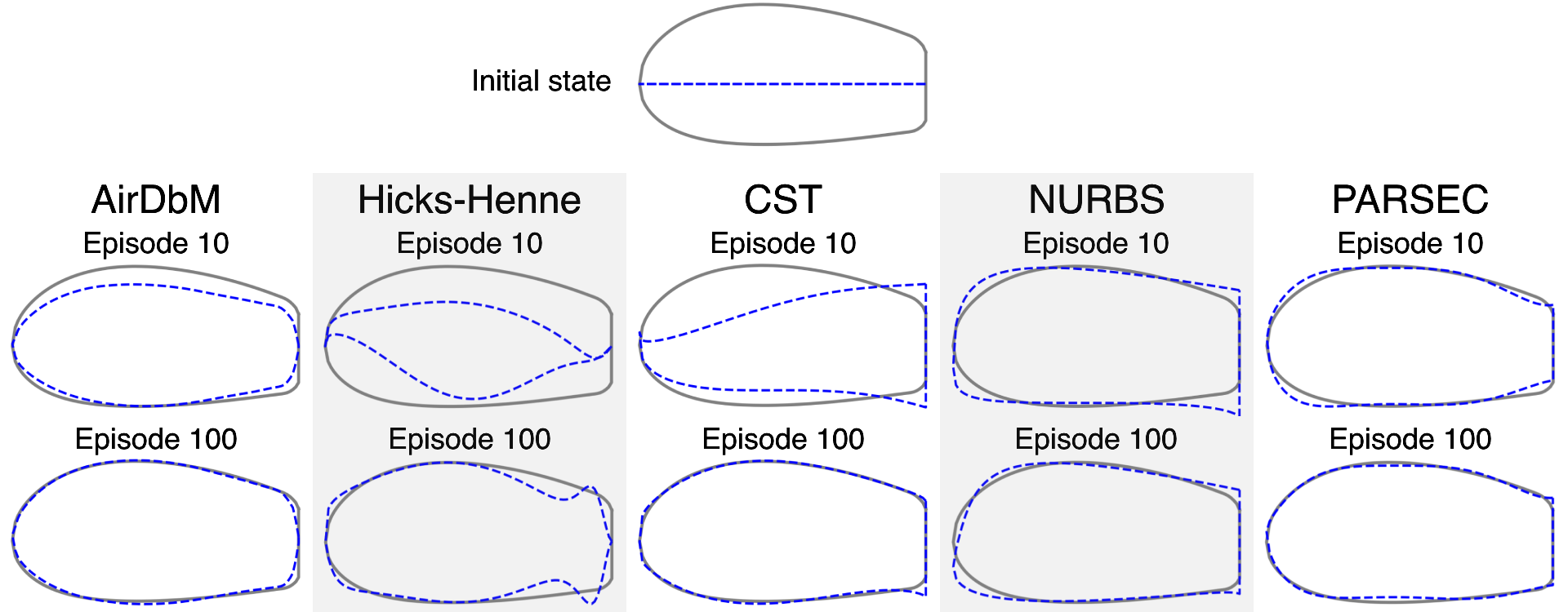}
        \captionsetup{justification=centering}\caption{Althaus AH 93-W-480B}
        \label{fig:AirfoilGenGeom-b}
    \end{subfigure}
    \caption{Comparison of geometry generation for (A) NACA 2412 and (B) Althaus AH-W-480B target shapes using various airfoil shape generators. Starting from the same initial state, the guessed shapes (blue dashed line) after Episodes 10 and 100 are compared to the target airfoil (gray solid line) for AirDbM, Hicks-Henne, CST, NURBS, and PARSEC.}
    \label{fig:AirfoilGenGeom}
\end{figure}

The most important setup in this RL framework is that the agent is completely unaware of the internal process of airfoil generation, thus the agent lacks \textit{a priori} insight into it. To ensure complete isolation, the agent does not directly control the design variables (which might imply knowledge of the airfoil generation method). Instead, it only manipulates normalized control inputs, like tuning `knobs' ranging from zero to one, that are linearly scaled to the design variables' bounds. By maintaining a consistent learning policy, we can then replace the airfoil parameterization method (airfoil shape generator) and assess the learning rate --- how quickly the agent's guesses converge towards the target --- over successive iterations (i.e., cumulative episodes). 

It is noteworthy that exploring different RL approaches is beyond the scope of this study. We use the \texttt{Gymnasium} framework \parencite{towers2024}, training a proximal policy optimization (PPO) agent with a multi-layer perceptron (MLP) surrogate policy \parencite{Raffin2021}, optimizing a reward signal defined as the negatively signed MAE of the guessed airfoil shape against a target one provided in the environment. For details regarding the airfoil parameterization methods used for comparison, adhering to 12 or 13 design variables --- AirDbM, Hicks–Henne, CST, NURBS, and PARSEC --- and the RL setup, refer to Appendix \ref{app:AP} and Appendix \ref{app:AG}, respectively.

In Figure \ref{fig:AirfoilGenGeom}, two representative outcomes of the airfoil geometry generation `game' are depicted: (A) NACA 2412 (thin airfoil) and (B) Althaus AH 93-W-480B (thick airfoil). Particularly, the latter case is one of the baselines of AirDbM, which is depicted to demonstrate the agent's lack of \textit{a priori} insight into the generation process as intended --- had the agent been aware of this fact, it could have arrived at the target directly. During each episode, the agent makes 100 attempts, iteratively refining its guess based on the best outcome from previous cumulative episodes, which gradually improves its control input tuning and ultimately results in better predictions (comparing Episode 10 to Episode 100).

\begin{figure}[tb!]
    \centering
    \includegraphics[width=0.8\linewidth]{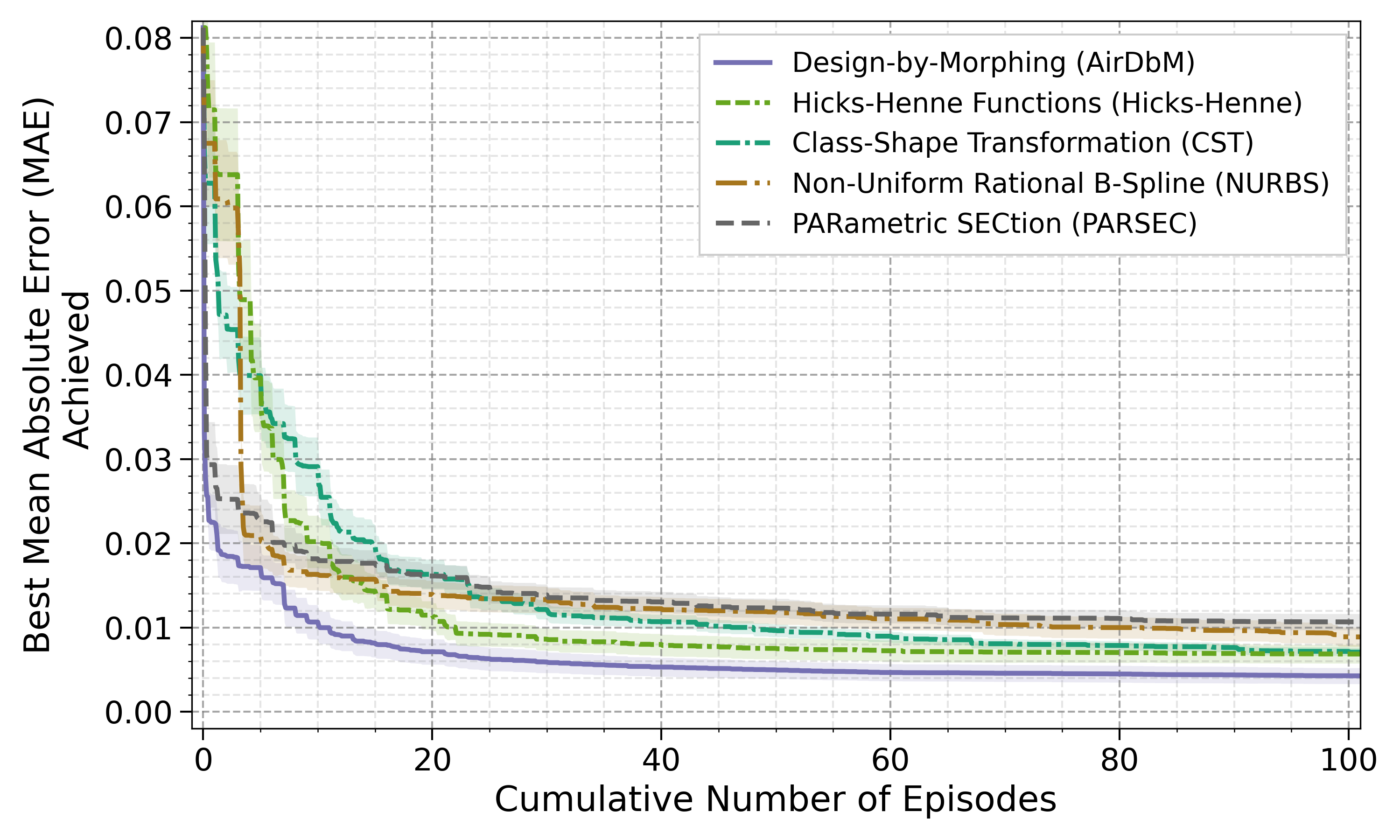}
    \caption{Quantitative evaluation of 5 airfoil shape generation methods --- AirDbM (the present Design-by-Morphing), Hicks-Henne, CST, NURBS and PARSEC --- illustrating average (solid line) and $ \pm 0.25\times$standard deviation (shaded area) of best mean absolute error (MAE) achieved over cumulative episodes across all 1,644 target airfoil shapes tested.}
    \label{fig:AirfoilGenComp}
\end{figure}

Compared against the other four conventional airfoil parameterization methods, AirDbM exhibits relatively fast convergence to the target in these two representative cases. For instance, looking into Episode 10 of the Althaus AH 93-W-480B environment (Figure \ref{fig:AirfoilGenGeom-b}), AirDbM's guess, albeit slightly thin yet, already becomes akin to the target airfoil shape, while the other guesses either are still far from airfoil shapes (Hicks-Henne and CST) or suffer from bloated leading edge curvature (NURBS and PARSEC). Despite the agent's unawareness, AirDbM inherently possesses the feature information of airfoils in the baselines. Therefore, for any weight inputs, the resulting shape is likely to be an airfoil shape as it is constructed by the mixture of the existing design features. In this regard, PARSEC, which more explicitly carries airfoil design features (since the design variables are directly geometric parameters of airfoils), is expected to show fast convergence but it is presumably the method's fundamental inferiority in reconstructing airfoil shapes that limits the performance \parencite[see][p. 1447]{Sheikh2023}.

For quantitative and non-prejudiced evaluation, all 1,644 target airfoil shapes in the database were tested under the same learning setup. A comprehensive comparison result is shown in Figure \ref{fig:AirfoilGenComp}. The trends of the best MAE measures achieved over cumulative episodes are plotted with respect to the five airfoil shape generation methods under consideration, where the solid line is the average of the entire 1,644 environment runs at each episode while the shaded area represents $\pm 0.25 \times$ standard deviation (the factor of 0.25 is merely for visual clarity to minimize overlapping between the shaded areas). 

In line with the findings from the representative cases, AirDbM overall exhibits the fastest decrease in the best MAE for initial episodes, keeping the lead up to the long run (Episode 100). PARSEC initially shows comparable decreasing rate in MAE with AirDbM, but as the episode accumulates, the performance gap widens, ultimately remaining the worst performance. The other three methods, Hicks-Henne, CST and NURBS, show relatively decent decrease, but their performances gradually improve and in the long run all arrive in between PARSEC and AirDbM.

Based on these results, it can be concluded that AirDbM, using just 12 systematically selected baseline airfoils, not only matches the reconstruction accuracy of conventional parameterizations but also excels in adaptability and learning efficiency when integrated with RL agents. In a full comparison across 1,644 target airfoils in the database, AirDbM enabled the agent --– as an unbiased designer of no prior knowledge --– to achieve lower mean absolute error and faster convergence than all compared airfoil parameterization methods, maintaining the same or a larger number of design variables. 

\section{Discussion}\label{discussion}
The present study has successfully demonstrated an improved Design-by-Morphing (DbM) approach, AirDbM, which significantly reduces design-space dimensionality for airfoil design and optimization. By focusing on maintaining design diversity through effective reconstruction of the rich airfoil database, our study achieved a substantial reduction in the number of baseline shapes required. The resulting AirDbM approach not only yielded benefits in multi-objective aerodynamic optimization, such as accelerated convergence and even partial enhancement of the Pareto-optimal solutions, as demonstrated in our former example (\S \ref{airmultiopt}), but also showed excellence in airfoil shape generation compared to several conventional parameterization methods.

Nonetheless, it should be admitted that the dimensionality reduction in AirDbM primarily concentrated on the \textit{geometric} feature preservation. Although this approach ensures broad geometric coverage, the aerodynamic optimization results suggest a trade-off, as exemplified by the inability to reach the extreme stall tolerance values achieved previously with a larger baseline set. Such a geometrically-focused compact design space appears to limit the exploration of more aerodynamically diverse or specialized design candidates.

Thus, future developments of the DbM framework for airfoil design could possibly benefit from incorporating aerodynamic considerations more directly into the baseline selection process. Beyond geometric diversity, selecting or augmenting baseline sets with airfoils known for specific, superior aerodynamic characteristics (e.g., high $C_l$ and low $C_d$) could allow the design space to better support the exploration of dynamically advanced airfoil designs. In particular, considering the reconstruction rate convergence observed (see Figure \ref{fig:ConvergenceTrend}), one may only choose the first 10 or 11 airfoils from the current baseline set and supplement the remaining slots with designs proven for their aerodynamic merit, creating a \textit{hybrid} baseline set. This can also be supplemented by additional dimensionality reduction efforts that preserve the essential design space scope, such as employing $n$-sphere coordinate variables for $(n+1)$ weight factor mapping, or pruning the design space by leveraging internal problem symmetries \parencite[e.g.,][]{Lee2024}.

It is clear that the significant design-space dimensionality reduction achieved by AirDbM mitigates the \textit{curse of dimensionality}. The reduction in the number of design variables is crucial as it opens up possibilities for integrating more computationally intensive or higher-fidelity solvers into the efficacious optimization loop. The current reliance on inexpensive solvers like XFOIL 6.99 facilitates rapid design exploration, but could be replaced or augmented by, for example, Reynolds-averaged Navier-Stokes (RANS) simulations or even experiments. In this way, we could pave the way for optimizing more realistic, 3D wing designs or tackling more complex aerodynamic phenomena necessitating higher fidelity. Such an approach requires efficient optimization algorithms with smaller data points of exploration rather than large-sample algorithms like GA, such as those based on Bayesian inference \parencite[e.g.,][]{Sheikh2022} or the use of PPO agents explored in our latter example (\S \ref{airfoilgeomgen}), for sample efficiency.

The data-efficient, interpretable parameterization of AirDbM (and DbM more broadly) reveals significant implications for machine learning-driven design. While deep generative models like generative adversarial networks (GANs) \parencite[e.g.,][]{Chen2020,Wang2023,Xie2024} excel at synthesizing novel designs through data-driven pattern recognition, they typically require thousands of training samples and lack inherent physical constraints. DbM can address these issues by providing geometrically consistent priors through systematic morphing of a baseline set containing $O(10^1)$ or perhaps fewer elements, generating physically plausible candidate designs that can seed and constrain GAN training. This symbiotic relationship enables generative models to focus on refining physical meaningful variations in avoidance of suffering from hallucinations of non-feasible geometry generations, leading to reduced training time while maintaining design feasibility. {\color{black} To sum up, while recent machine learning-driven methods (e.g., GANs and variational autoencoders, VAEs) reduce latent-space dimensionality when large datasets are available, DbM begins with a few known baseline designs to span a wide and physically relevant design space. These approaches imply their synergistic relation: DbM is not a direct competitor to these methods, but could rather be complementary.}

\section{Conclusions}\label{conclusions}
We addressed the challenge of reducing design-space dimensionality in Design-by-Morphing (DbM) for airfoil optimization by introducing AirDbM, an DbM-based airfoil design approach with a systematically reduced baseline set. Utilizing an effective forward search strategy, we identified a compact yet highly representative set of 12 baseline airfoils selected from the UIUC database of 1,644 airfoils. This reduced set retained broad airfoil design capability, as demonstrated in reconstruction tests where 98 \% of the database was reproduced within a mean absolute error of 0.005. This performance rivals --- and, in terms of dimensionality, surpasses --- the previous DbM efforts that used 25 baselines, thereby achieving a substantial reduction in design parameters without compromising geometric diversity.

The efficacy of AirDbM was quantitatively demonstrated in both multi-objective aerodynamic optimization using a genetic algorithm (GA) and airfoil geometry generation in the context of reinforcement learning. In aerodynamic shape optimization aimed at maximizing both lift-to-drag ratio and stall tolerance, AirDbM achieved accelerated convergence. Its hypervolume indicator value surpassed that of the earlier 25-baseline study in significantly fewer GA generations. The resulting Pareto front identified new Pareto-optimal solutions with enhanced lift-to-drag ratios, especially at low to moderate stall tolerances. In a comparative study that employs machine agents as unbiased designers from a reinforcement learning framework, AirDbM achieved faster convergence and lower errors than several conventional airfoil parameterization methods, while using a similar number of design variables.

These findings lay the groundwork for further advancements in the DbM methodology for airfoil design and optimization. Future directions may include incorporating aerodynamic performance criteria into the baseline selection process to create hybrid sets that maintain geometric representativeness while targeting specific aerodynamic objectives. Additionally, the computational efficiency gained from operating in a lower-dimensional design space facilitates the integration of higher-fidelity solvers, paving the way for a transition from 2D airfoil analysis to more realistic 3D wing design applications. Such developments are anticipated to be synergistically combined with modern machine learning-driven generative design approaches for expedited optimization.


\section*{Conflicts of Interest} 
The authors declare no conflict of interest.

\section*{Author Contributions}
\noindent\textbf{Sangjoon Lee}: Formal analysis, Investigation, Methodology, Software, Writing-original draft. \textbf{Haris Moazam Sheikh}: Conceptualization, Methodology, Project administration, Supervision, Writing-review \& editing.

\section*{Funding}
This work used Anvil at Purdue RCAC through allocation PHY250071 from the Advanced Cyberinfrastructure Coordination Ecosystem: Services \& Support (ACCESS) program, which is supported by U.S. National Science Foundation grants \#2138259, \#2138286, \#2138307, \#2137603, and \#2138296.

\section*{Data Availability}
The present Design-by-Morphing (AirDbM) codes and data, incorporated into the multi-objective airfoil optimization problem in this study, is available in the public repository on \href{https://github.com/UCBCFD/DbMAirfoilOpt}{https://github.com/UCBCFD/DbMAirfoilOpt}. Additional codes or data can be shared on a reasonable request to the corresponding author.

\section*{Acknowledgments}
A significant portion of the computations for data analyses was performed on the Yellowstone at the Stanford HPC Center, supported through awards from Intel, National Science Foundation, DOD HPCMP, and Office of Naval Research.


\printbibliography


\renewcommand\theequation{\Alph{section}\arabic{equation}} 
\counterwithin*{equation}{section} 
\renewcommand\thefigure{\Alph{section}\arabic{figure}} 
\counterwithin*{figure}{section} 
\renewcommand\thetable{\Alph{section}\arabic{table}} 
\counterwithin*{table}{section} 

\begin{appendices}

\section*{Appendix}

\section{Airfoil Design Methods}\label{app:AP}
\begin{table}[tb!]
\footnotesize
  \caption{Airfoil parameterization methods considered for comparison}
  \label{tab:appA}
  \begin{tblr}{
      colspec = {c l X[c] X[c]},
      rows={m, abovesep=8pt, belowsep=4pt},
      row{even} = {lightgray!10},
      row{1} = {rowsep = 2pt},
      width = \linewidth,
    }
    \hline \hline
    \textbf{Method} & \textbf{Design Variables (DVs)} & \textbf{Remarks} \\
    \hline \hline
    AirDbM & 
    $w_i$: Morphing weight factors $\in [-1.0, 1.0]$ ($i=1,~\cdots,~12$) & 
    See Table \ref{tab:baseline_airfoils} for the baselines \\
    Hicks-Henne &
    \makecell[l]{ $p_{u,i}$: Upper bump powers $\in [1.0, 4.0]$ ($i=1,~2,~3$) \\
    $a_{u,i}$: Upper bump amplitudes $\in [-0.2, 0.2]$ ($i=1,~2,~3$) \\ 
    $p_{l,i}$: Lower bump powers $\in [1.0, 4.0]$ ($i=1,~2,~3$) \\ 
    $a_{l,i}$: Lower bump amplitudes $\in [-0.2, 0.2]$ ($i=1,~2,~3$) } & 
    Base: flat plate; cosine-distributed bump points; see \textcite[][]{Hicks1978} \\
    CST & 
    \makecell[l]{ $N_1$: 1st class function exponent $\in (0.0,2.0]$ \\ 
    $N_2$: 2nd class function exponent $\in (0.0,2.0]$ \\
    $A_{u,i}$: Upper Bernstein coefficients $\in [-0.5, 0.5]$ ($i=1,~\cdots,~4$) \\
    $\Delta \xi_u$: Upper trailing edge height $\in [-0.5, 0.5]$ \\
    $A_{l,i}$: Lower Bernstein coefficients $\in [-0.5, 0.5]$ ($i=1,~\cdots,~4$) \\
    $\Delta \xi_l$: Lower trailing edge height $\in [-0.5, 0.5]$ } &
    See \textcite[][]{Kulfan2006} \\
    NURBS & 
    \makecell[l]{ $x_1$: 1st control point's $x$-coordinate $\in [0.0,1.0]$ \\
    $y_1$: 1st control point's $y$-coordinate $\in [-0.5,0.5]$ \\
    $x_2$: 2nd control point's $x$-coordinate $\in [-0.5,0.5]$ \\
    $y_2$: 2nd control point's $y$-coordinate $\in [-0.5,0.5]$ \\
    $x_3$: 3rd control point's $x$-coordinate $\in [0.0,1.0]$ \\
    $y_3$: 3rd control point's $y$-coordinate $\in [-0.5,0.5]$ \\
    $y_{te,u}$: Upper trailing edge height $\in [-0.5, 0.5]$ \\
    $y_{te,l}$: Lower trailing edge height $\in [-0.5, 0.5]$ \\
    $\omega_i$: Control point weights $\in [0.1, 5.0]$ ($i=1,~\cdots,~5$) } &
    3rd-order B-spline with evenly distributed knots; see \textcite{Piegl1997} \\
    PARSEC & 
    \makecell[l]{ $r_{le,u}$: Upper leading edge radius $\in [0.0, 1.0]$ \\
    $x_{u}$: Upper crest's $x$-coordinate $\in (0.0, 1.0)$ \\
    $y_{u}$: Upper crest's $y$-coordinate $\in [-0.5, 0.5]$ \\
    $y_{xx,u}$: Upper crest curvature $\in [-0.5, 0.5]$ \\
    $r_{le,l}$: Lower leading edge radius $\in [0.0, 1.0]$ \\
    $x_{l}$: Lower crest's $x$-coordinate $\in (0.0, 1.0)$ \\
    $y_{l}$: Lower crest's $y$-coordinate $\in [-0.5, 0.5]$ \\
    $y_{xx,l}$: Lower crest curvature $\in [-0.5, 0.5]$ \\
    $y_{te}$: Trailing edge mid-position $\in [-0.5, 0.5]$ \\
    $t_{te}$: Trailing edge thickness $\in [0.0, 1.0]$ \\
    $\alpha_{te}$: Trailing edge direction $\in [-\pi/4, \pi/4]$ \\
    $\beta_{te}$: Trailing edge wedge angle $\in [0, \pi/2]$} &
    See \textcite{Sobieczky1999} \\
    \hline \hline
  \end{tblr}
\end{table}

Throughout this work, we considered five distinct airfoil parameterization methods for comparison: AirDbM, Hicks-Henne, CST, NURBS, and PARSEC, as detailed in Table \ref{tab:appA}. A key aspect of this setup was the standardization of the dimensionality across these methods. The DbM, Hicks-Henne, CST, and PARSEC methods are configured to utilize 12 design variables. The NURBS parameterization was slight exception, making use of 13 design variables; this number was chosen to maintain the method's parametric integrity while aligning it as closely as possible with the 12-design variable target used by the other methods. This consistent dimensionality facilitated a fair comparison of the different methods' capabilities in airfoil geometry generation.

\section{Multi-Objective Airfoil Optimization Setup}\label{app:MO}
As detailed in \S \ref{airmultiopt}, the multi-objective airfoil optimization was conducted using the \texttt{gamultiobj} optimizer in MATLAB \parencite{MathWorksGamultiobj}, based on NSGA-II. The optimization aimed to identify superior airfoil designs, each parameterized by 12 morphing weight variables founded upon AirDbM that range from -1 to 1, with respect to lift-to-drag ratio as a primary design point and stall tolerance as a robustness for off-design operation.

\begin{table}[tb!]
\footnotesize
  \caption{Details of the multi-objective genetic algorithm used in this study}
  \label{tab:appB}
  \begin{tblr}{
      colspec = {X[c] X[l]},
      rows={m, abovesep=8pt, belowsep=4pt},
      row{even} = {lightgray!10},
      row{1} = {rowsep = 2pt},
      width = \linewidth,
    }
    \hline \hline
    \textbf{Option} & \textbf{Selection} \\
    \hline \hline
    Population size & 372 \\
    Total generations & 1,000 \\
    Selection scheme & Binary tournament (Pareto fraction = 0.35) \\
    Crossover scheme & Intermediate crossover (Crossover fraction = 0.8) \\
    Mutation scheme & Adaptive feasible \\
    Distance measure of individuals & Crowding distance in fitness function space \\
    \hline \hline
  \end{tblr}
\end{table}

The optimizer was configured with a population size of 372 individuals and a maximum of 1,000 generations. The evolution of this population was driven by the following genetic operators: selection based on a tournament approach considering non-domination rank and crowding distance (calculated in the objective or fitness function space), an intermediate crossover strategy with a crossover fraction of 0.8, and an adaptive feasible mutation scheme that introduces variations using randomly generated directions adapting to the previous generation. Following the initialization of \textcite{Sheikh2023}, the initial population was composed by incorporating outcomes from two preliminary single-objective GA runs (each with a population of 128 and run for 100 generations) for each of the design targets --- lift-to-drag ratio and stall tolerance. The remaining individuals of the initial population were randomly distributed. A summary of these algorithmic parameters is provided in Table \ref{tab:appB}.

At each generation, the XFOIL performance evaluations were parallelized; this study utilized up to 128 cores to perform XFOIL analyses concurrently for 128 airfoil samples, significantly reducing the overall duration of the optimization process.

\section{Airfoil Geometry Learning Setup}\label{app:AG}
In \S \ref{airfoilgeomgen}, the airfoil geometry generation task was formulated as a reinforcement learning problem and addressed using a proximal policy optimization (PPO) agent. The PPO agent utilized a multi-layer perceptron for both the actor and critic networks. Key hyperparameters for the PPO algorithm included the learning rate 
of 0.0003, the number of steps per update
of 2048, the batch size 
of 64, the epochs per update 
of 10, the discount factor
of 0.99, the generalized advantage estimator (GAE) lambda 
of 0.95, the PPO clipping range 
of 0.2, the entropy coefficient 
of 0, and the value function coefficient 
of 0.5. Other PPO parameters largely followed the default values as in the Stable-Baselines3 library (v2.6.0) \parencite{antonin2025}.

The learning environment was configured for episodic tasks. Each episode consisted of 100 steps. The observation space provided to the agent was the set of 12 or 13 normalized inputs ranging from $0.0$ to $1.0$. The reward at each step was calculated as the negative of the MAE, incentivizing the agent to produce airfoils more closely matching the target. Each PPO model was trained for a total of 10,000 steps (i.e., 100 episodes).

\end{appendices}

\end{document}